\documentclass[letterpaper, 10 pt, conference]{IEEEtran}  % Comment this line out if you need a4paper

\usepackage{lmodern}
\usepackage[T1]{fontenc}

\IEEEoverridecommandlockouts                              
\usepackage{graphics} % for pdf, bitmapped graphics files
\usepackage{epsfig} % for postscript graphics files
\usepackage{mathptmx} 
\usepackage{times}
\usepackage{amsmath} 
\usepackage{amssymb} 
\usepackage{xcolor} % for text colors
\usepackage{subcaption}
\usepackage{wrapfig}
\usepackage{algorithm}
\usepackage{algpseudocode}
\usepackage{url}
\usepackage{caption}
\usepackage{subcaption}
\usepackage{epstopdf}
\usepackage{multicol}
\usepackage{multirow}
\usepackage{times}
\usepackage{colortbl}
\usepackage{array}
\usepackage{booktabs} % for nicer tables

\title{\LARGE \bf
Multi-Robot Box Transport over Different Surfaces with \\Decentralized Role-based Proportional Control
}

\author{Aditya Bhatt$^{1}$, Himavarshini Yarragangu$^{1, *}$, Urvish Shah$^{1, *}$, \\ Venkata Sai Yaswanth Mohan Thota$^{1, *}$ and Souma Chowdhury$^\dagger$ % <-this % stops a space
\thanks{$^{*}$ Equal Contribution}
\thanks{$^\dagger$ Corresponding Author, soumacho@buffalo.edu}
\thanks{$^{1}$ Mechanical \& Aerospace Eng.,
        University at Buffalo, Buffalo, NY 
        }%
\thanks{This work was supported by the NSF award CMMI 2048020.
Any opinions, findings, conclusions, or recommendations expressed in this paper are those of the authors and do not necessarily reflect the views of NSF. Facilities support from the University at Buffalo's Center for Embodied Autonomy and Robotics (CEAR) is also acknowledged. }
\thanks{This work is accepted to be presented at the 2026 IDETC-CIE.}
}

\begin{document}

\maketitle

\begin{abstract}
Collaborative transport of objects via pushing by multiple robots has many applications, ranging from construction and warehouse environments to post disaster debris clean-up. Achieving collaborative transport over surfaces with different inclination and friction properties however poses unique challenges. To address these challenges, this paper presents an asynchronous decentralized task and motion planning approach for transporting rectangular boxes of varying mass over flat, uphill and downhill terrain. Such a decentralized approach alleviates communication, synchronization and consensus needs and mitigates single point of failure issues. Our approach, called \textbf{R2P2} or Roles with Rules and Proportional-control Primitive, assigns roles (e.g., \textit{push}, \textit{support} and \textit{prevent}) to robots based on rules cognizant of the mode of manipulation needed (box rotation vs translation); this is followed by either rule-based control or proportional control of robot velocity based on the roles. Each robot is assumed to observe the location and heading of self and the box in executing the role and controls. R2P2 is evaluated with a six-robot team deployed in a simulator built using NVIDIA IsaacSim -- demonstrating generalizability across different surface friction/inclination and box mass scenarios, and better success rate compared to a standard virtual-leader-follower method. R2P2 is also successfully validated with a physical experiment, where it is executed onboard four turtlebots tasked with moving a 1.2 kg box. 
\end{abstract}

% \footnotetext[$\dagger$]{These authors contributed equally to this work.}
%%%%%%%%%  NOMENCLATURE (OPTIONAL) %%%%%%%%%%%%%%%%%%%%%%%%%%%%%%%%%
%%
%% To change space between the symbols and  definitions, use \begin{nomenclature}[Xcm] where X is a number 
%% The unit cm can be replaced by any LaTeX unit of dimension: pt, in, ex, em, pc, etc.
%% Default is 2em.
%%
%% \EntryHeading{..} produces an italicized subheading in the nomenclature list, e.g., \EntryHeading{Greek letters}

%%%%%%%%%  BODY OF PAPER %%%%%%%%%%%%%%%%%%%%%%%%%%%%%%%%%

\section{Introduction}  \label{sec:intro}
Managing warehouse logistics, transportation of materials on construction sites, space/extraterrestrial exploration, and removal of debris from disaster and remote sites are examples of applications that can potentially benefit from capabilities of autonomous planar box manipulation or box transport jointly carried out by multiple ground robots \cite{li2022survey, rosenfeld2016human, petersen2019review}. 
Such a collaborative object transport approach with a team of robots offers major advantages in terms of size of objects that could be transported, and associated capacity and operational efficiency. For example, an individual robot may not even have the capacity (push-force, endurance etc.) to carry out the transport; and with a team of robots, the transportation time can greatly decrease. Within this domain of collective object transport, this paper particularly focuses on non-prehensile transport, where grasping and gripping of the object is not involved. Moreover, we assume non-holonomic wheeled robots, given their mainstream popularity in the related use cases. 
% In addition, the non-prehensile transport is beneficial 
Non-prehensile collective box transport over realistic surfaces or terrain, involving inclination and friction effects, remains a challenging problem; this is mainly because multi-robot co-ordination must be achieved considering contact dynamics at three fronts: 
% box transport using MRS is a challenging problem from the theoretical perspective because the co-ordination has to consider contact dynamics at three fronts, 
i) contact between the robot and the box, ii) contact between the robot and the terrain and iii) contact between the box and the terrain. 
The goal of this paper is to develop a computationally lightweight task and motion planning method that can be implemented in a decentralized manner across (onboard) a team of robots to achieve robust box transport over flat and inclined surfaces (up-hill and down-hill) with different friction property. In the rest of this section, we summarize the effects of terrain to be cognizant of with regards to achieving successful (collective) box transport, followed by overview of the associated planning challenges, existing approaches to solve them, converging on our proposed concept and contributions of this paper. 

Given the three contact fronts, the inclination and friction properties of the operational surface has the following major effects that must be accounted for to achieve successful collaborative box transport. Firstly on flat rough terrain, adequate force is required to overcome the friction force on the box, and in a manner that does not cause undesirable rotation of the box. On uphill terrain, adequate force is required to move the box against gravity and friction forces, while also being supported to avoid undesirable rotation and slipping out (and down) of the box. On downhill terrain, adequate support is required to ensure that the box does not go into uncontrolled downward sliding (and rotation).

The problem of box transport in general requires decision making at three levels. 
{\textbf{1)}} Optimal waypoint planning for the box, 
{\textbf{2)}} Robot formation planning, 
{\textbf{3)}} Robot motion control. 
In an environment with known static obstacles, the first step is to plan a time optimal or energy optimal path of the box in terms of sequence of waypoints. 
The path planning algorithms, either sampling based (e.g, RRT) or optimization based (e.g, A$^{*}$, Djikstra) require a cost function to compute the cost of the path to the candidate waypoint. The objectives of the cost function (e.g. energy) can be obtained by achieving the box transport to that candidate waypoint. This leads to solving the next two sub-problems of robot formation planning and motion control. 
Prior work on robot formation planning has focused on deriving formulations that either use force analysis (contact normals) or caging (geometric box properties) to arrive at a robot formation \cite{ebel2022finding, wan2017multirobot}. Motion control formulations then optimize energy (e.g minimize control velocities, re-formations) while adhering to the constraints (more details in Sec.~\ref{sec:rel-work}). 

Robot formation and motion control has also been solved by formulating it as a joint task and motion planning (TAMP) problem.
In non-prehensile object manipulation, task planning is usually attributed to searching for the contact points for the robot-team and motion planning corresponds to force computations \cite{tang2024collaborative, ebel2022finding, moon2012cooperative}. The resultant formulation is usually a centralized one and not amenable for on-board computation. Moreover, these approaches may not readily extend to inclined terrains (more details in Sec.~\ref{sec:rel-work})
% In addition, it may also limit the generalizability with respect to the type of the terrain.  
% \textcolor{red}{You need 2 sentences here to summarize why these methods are not good enough and what's the rationale for your alternate approach. (and point to Sec. II for details)}

In contrast, in this paper we explore a new {asynchronous \textit{decentralized}} TAMP formulation. 
% Decentralized decision-making has two type of advantages. First, hardware oriented. The centralized systems risk mission compromise with the failure of the decision-making system. Second, a decentralized framework offers better scalability. The number of decision variables in centralized framework scales with the increase in the number of robots. With an increase in the decision variables, the complexity of the decision-making increases. The main challenge in decentralized decision-making is the limited on-board computing. 
Decentralized decision-making systems eliminate the single point of failure and over-reliance on communication with a central node associated with the centralized systems. It also promotes scalability with the number of robots, since the size of the decision space for planning does not necessarily grow with the number of robots. In such decentralized planning systems, the decision is computed based either on the local (self) information or by integrating it with the peers' information (communicated by peers). Being asynchronous further alleviates the needs for synchronizing plans and achieving consensus. In planning problems (e.g. Multi-Robot Task Allocation (MRTA) \cite{gerkey2004formal, ghassemi2018decentralized} or Multi-Robot Signal-Search \cite{ghassemi2022penalized, bhatt2025learning}) the time-horizon of planning instances allow for decision-making frameworks to rely on complete information of peer-states. In contrast, in control related high-frequency decision-making problems, using complete/recent information of the peer states may not be viable, leading to suboptimal or infeasible decisions because of communication dependencies \cite{yamchi2017distributed}. 
% In case of collaborative transport applications, one of the main challenges is the limited observations. This is because the control frequency is much higher compared to the planning frequency.  
% Decentralized decision-making systems offer mission-efficiency, fault-tolerance and scalability \cite{ghassemi2018decentralized, parker2016multiple}, with the main challenge being computing the decisions with limited compute power. 
% \textcolor{red}{Expand upon the advantages of decentralized decision-making, followed by challenges, all with citations}
To address this issue, here we associate each robot's \textit{task decision} with its ``\textit{role assignment}''; where, based on the robot's role, its own state and state of the box, the motion planner computes its velocity control. 
% The complete optimization problem, which includes optimization of the box path along with task decision and motion control, requires dynamic model of the environment and may result in optimization over simulation. 
% In its entirety, the problem statement at hand requires optimization of intermediate waypoints along with the TAMP variables \cite{tang2024collaborative}. It requires dynamic model of the environment and often results in optimization over simulation. 
% In this paper, we do not solve the problem in its entirety. 
% In this paper, we propose a 
% Instead, in this paper, we propose a compute-efficient, decentralized and scalable approach to solve this problem.
This formulation is posited to provide a compute-efficient and decentralized approach to achieving the collaborative box transport, with the following assumptions: i) The intermediate waypoints through which the box needs to be taken to the goal point are already available. Therefore, it is also implicit that no two waypoints have obstacles between them. 
In other words, this assumption corresponds to availability of the collision-free reference path for the box transport. This has been a common assumption in the prior work \cite{ebel2024cooperative, wan2017multirobot}. The collision-free reference path can be computed using a path-planning algorithm (A$^{*}$, Djikstra or RRT), by modeling the box and the robots with a convex polygon envelop. 
ii) The longest length of the box is greater than twice the diameter of the robots. In other words, the longest box edge should allow placement of two robots over the edge. 
iii) The initial relative robot positions have been pre-computed, since it can be readily accomplished with methods such as optimality criteria \cite{ebel2022finding}, caging criteria \cite{wan2017multirobot, fink2008multi} or  heuristics. iv) Each robot has perfect observation of the following: self-position, self-heading, box-position, and box-heading, all in global coordinate system, and stop flag from each peer robot $j$ (a boolean parameter to request other robots, $i\ne j$, to stop moving). 
%  The fourth assumption is related to   The edge length of the box is at least twice greater than the radius of the The assumption requires 
% The assumption requires the longest length of box greater than twice the radius of the robot

% \textcolor{red}{CLEARLY state the assumed observation space of each robot here} 
Given the initial robot positions and the assumed observation space of each robot, our decentralized framework allows each robot to self-assign the role and decide its velocity at each timestep over the mission horizon. 
In accordance with Occam's razor principle, we define a set of simple yet formal (intuitive) rules to allocate roles to the robots. 
Following that, the motion planner computes the velocities for the robot in accordance with its role. Contingent on the role, one of two velocity control approaches is used: a fuzzy rule based controller or a proportional controller. We call this new TAMP formulation and solution approach,  
\textbf{R2P2: Roles with Rules and Proportional-control Primitive}. In its current form, R2P2 is a lightweight, compute-efficient decentralized decision-making framework, which uses heuristic rules to assign roles. These heuristics can however be optimized in the future or learnt for energy/time optimal performance or generalizability across scenarios. 
% Further discussion on this is in Sec.~\ref{subsec:discussion}.
% it has the scope of optimizing over the heuristics. 

% In this paper, the case-studies present the generalizability of the method with rectangular box. However, with small modifications in the 

\begin{figure}
    \centering
    \includegraphics[width=0.95\linewidth]{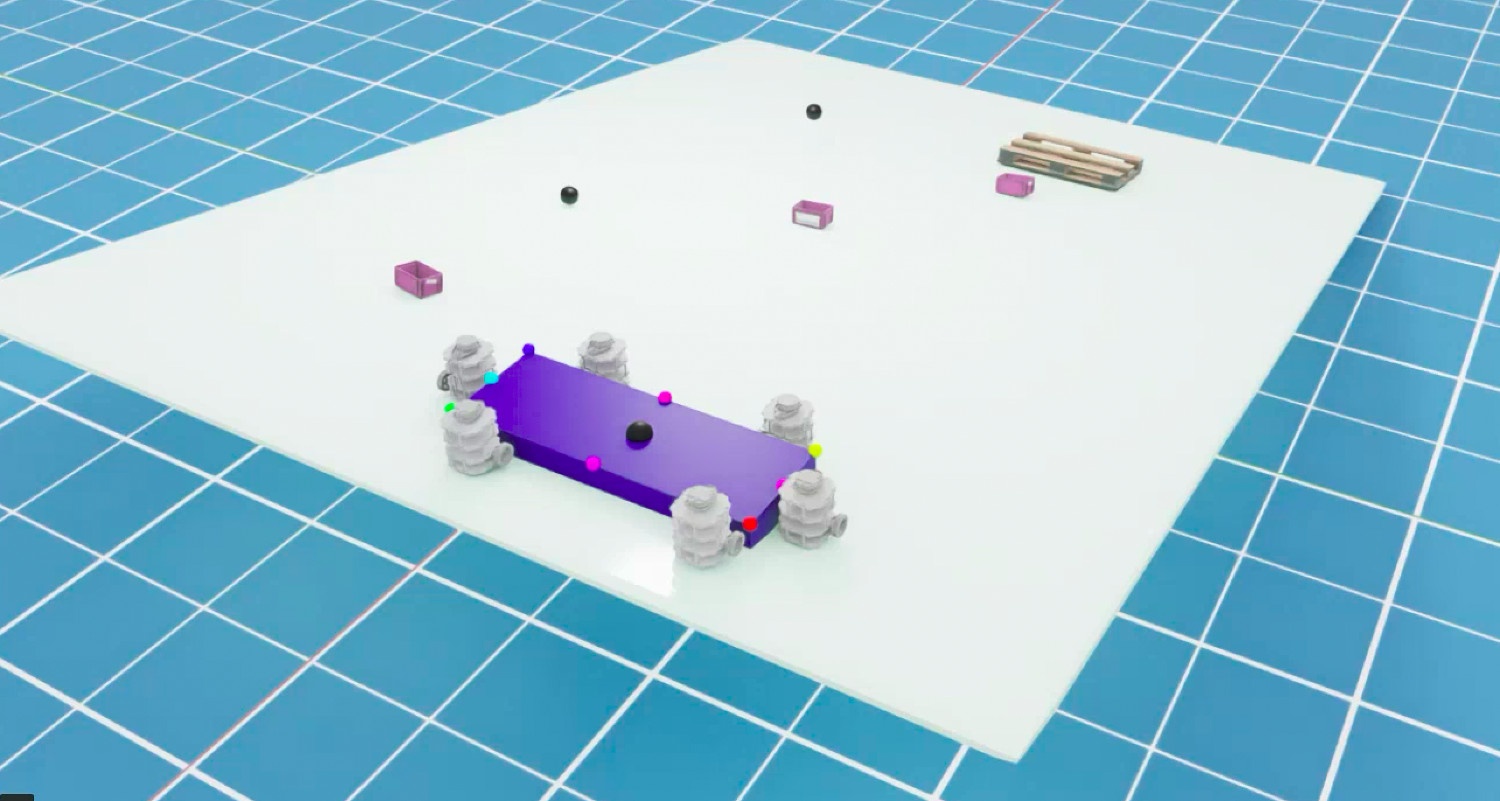}
    \caption{Collaborative Box transport using a team of turtlebots in NVIDIA IsaacSim. The goal is to transport the box (down-hill) to the goal-location through the waypoints (shown as spheres), on a plane at 5$^\circ$ incline. The goal location is the wooden plank shown down the plane.}
    % Adjust this value to your needs

    \label{fig:isaac}
\end{figure}

% \begin{figure}
%     \centering
%     \includegraphics[width=0.95\linewidth]{Figures/fig1_new_check.png}
%     \caption{Collaborative Box transport using a team of turtlebots in NVIDIA IsaacSim. The goal is to transport the box (down-hill) to the goal-location through the waypoints (shown as spheres), on a plane at 5$^\circ$ incline. The goal location is the wooden plank shown down the plane.}
%     % Adjust this value to your needs

%     \label{fig:isaac}
% \end{figure}

We implement and test R2P2 for rectangular box transport on a flat terrain and on inclined plane using a high fidelity simulation environment built with NVIDIA IsaacSim. Figure \ref{fig:isaac} shows a snapshot of a mission scenario in this simulator. The robots need to transport the box to the goal point while avoiding the obstacles, by pushing it through the intermediate waypoints. We test the generalizability of the method by varying the terrain friction coefficient and box mass. The team size is kept fixed at 6 for all simulation experiments. 
We also validate R2P2 with a physical experiment that uses 4 turtlebots to move a box in a lab-scale motion capture setup.

The key contributions of this paper can thus be summarized as:
\textbf{1)} A new abstraction of the collaborative box transport problem as a joint task and motion planning problem where role assignments define tasks and coherent velocity control defines motion planning.  
\textbf{2)} A compute-efficient, decentralized approach to solving this problem that uses  a situation-aware rule-based role assignment followed by rule-based or proportional control. 
% The role assignment is done based on a set of rules. A separate motion control is used for each role. 
\textbf{3)} Development and use of a high-fidelity simulation environment for extensive evaluations of the proposed method across diverse (including flat/inclined/smooth/rough) surfaces, and a physical lab-scale setup for experimental validation/demonstration. 
\textbf{4)} Use of a moderate-fidelity simulation environment based on Pybullet for analysis of the sensitivity of the mission completion time w.r.t.~the major heuristics in our R2P2 method, across environmental variations. 

The rest of the paper is structured as follows. Section ~\ref{sec:rel-work} describes the existing work in collaborative object transport. Then, we explain the associated problem formulation in Sec.~\ref{sec:pb-form}. Subsequently, in Sec.~\ref{sec:method}, we describe our proposed method. Following that, Sec.~\ref{sec:num-expts} presents the numerical experiments and ensuing analysis. In Sec.~\ref{sec:hardware}, we present the validation of R2P2 through lab-scale experiments. 
Finally, Sec.~\ref{sec:conc} gives our concluding remarks and future work. 

\section{Related Work} \label{sec:rel-work}

% The extensive literature review on the types of cooperative box transport can be found in \cite{tuci2018cooperative}. This topic has been of interest to the community since more than two decades. 
Object manipulation using Multi-Robot-Systems (MRS) has been a topic of interest to the MRS community for more than two decades. 
Among the three levels of decision making stated earlier, a lot work has gone into developing formulations for robot formation planning and motion control. The methods/formulations developed therein can be broadly classified into two categories: 1) Force analysis based methods and 2) Geometry based methods. Tuci et al. \cite{tuci2018cooperative} has extensively described the previous methods. Here, we will briefly discuss some of the new methods in both the categories. 

\textbf{Force analysis based methods.} Ebel el at in \cite{ebel2021non} used convex polytope hull formed by the contact normals to assess the robustness of the formation. The robustness formulation is used to allocate robot positions and a distributed Model Predictive Control (MPC) is used to assign velocity control for omni-directional robots. Later in \cite{ebel2022finding}, they propose a constrained optimization formulation to optimize the control cost for non-holonomic robots. Finally, they propose optimal control problem formulations and solve it using MPC \cite{ebel2024cooperative}. 
% in \cite{ebel2024cooperative}, they develop two optimal control problems. The first one guides the differential drive robot towards its desired relative object position without colliding with the object and the second maintains the formation. 
% Recently, Tang et al \cite{tang2024collaborative} proposed a hierarchical search method to optimize the object trajectory along with optimizing the contact points and the velocities. 
% These methods have been successfully tested over through physical experiments on flat terrains. 
% However, since these methods do not incorporate caging constraints, they may not work well on inclined terrains, especially with down-hill transport.  

Related solution approaches in this class of force analysis based methods can also be divided in terms of centralized and decentralized frameworks. Within centralized frameworks, Tang et al \cite{tang2024collaborative} propose real-time heirarchical optimization framework to compute the robot positions and the robot velocities. Jankowski et al \cite{jankowski2025robust} present an open-loop offline trajectory optimization framework.
Ebel at al \cite{ebel2022finding, ebel2024cooperative} propose a Distributed (Decentralized) Model-Predictive Control (DMPC) framework to compute the robot-positions and velocities. While they explicitly model the non-holonomic constraints and their framework handles different box shapes.
Since DMPC scheme involves optimization at each time-step, it is computationally expensive and may not be amenable for on-board compute. Even with the availability of on-board compute power, the optimization in its current form may not handle the dynamic uncertainties which require quick decision-making. 

\textbf{Geometry based methods.} These methods derive robot position formulations (usually in object configuration space) to ensure robust caging of the object. Wan et al \cite{wan2017multirobot} proposed formulations to derive the optimal number of robots along with their locations by formulating caging robustness. Subsequently, they use virtual leader-follower to compute velocity for omnidirectional robots. Vardharajan et al \cite{vardharajan2022collective} design an approach for caging based transportation of an object whose shape is not known apriori. 

The related work on box transport faces the following challenges:
1) Force analysis based methods do not incorporate the caging constraints, and hence they may not work on inclined terrains. 
2) The caging based strategies which have been tested on inclined terrains use omni-directional robots. 
In this work, we develop a decentralized control method for caging based box transport using differential drive robots with non-holonomic constraints. We test our framework on inclined planes with obstacles, as well as on flat terrains. 
As a baseline, we use the virtual leader follower method that has been commonly used in caging based box transport \cite{1308049, wan2017multirobot, tuci2018cooperative} and provides run-time cost similar to R2P2, as opposed to optimization based methods which generally have high compute cost.

\section{Problem Statement} \label{sec:pb-form}

%\vspace{-2mm}

\begin{figure}
    \centering
    \includegraphics[width=0.95\linewidth]{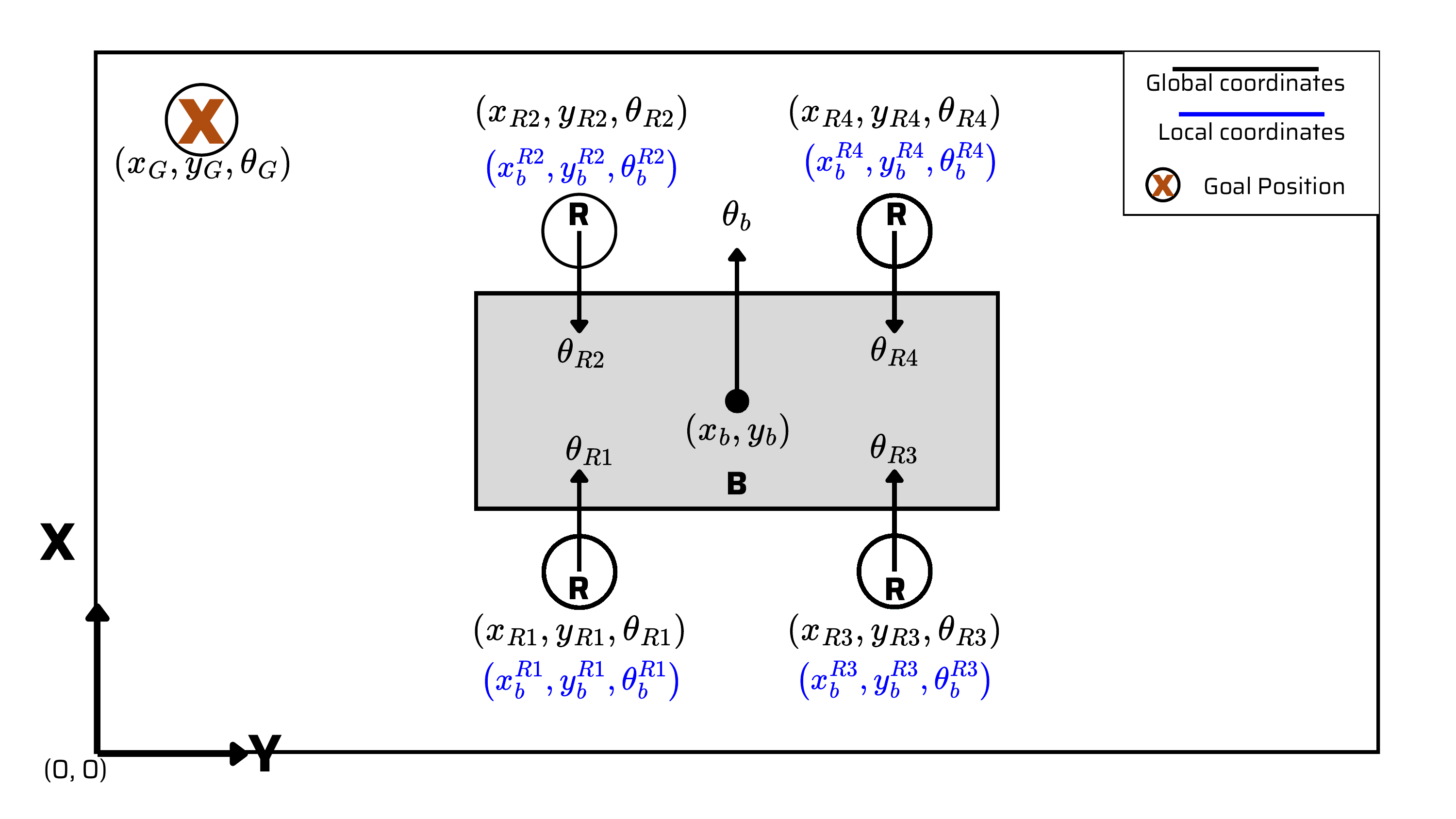}
    \caption{Global and local coordinate system of the object (represented by a rectangle) and the robots (represented by a circle). Goal is represented by X}
    %\vspace{-5mm} % Adjust this value to your needs

    \label{fig:global-local}
\end{figure}
% \begin{figure}
%     \centering
%     \includepdf[pages=-]{Figures/Global_and_local_corr_rob_obj.pdf}
%     \caption{Global and Local coordinates of Object and Robots}
% \end{figure}

This section explains the optimization problem formulation in its entirety. Note that in this paper, we do not perform the optimization, we just formulate the optimization problem. 
As shown in Fig.~\ref{fig:global-local}, let the initial global box position be represented as $b(0) =$ ($x_b, y_b$), the heading of the box as $\theta_b$, the global position of the robot $i$ as ($x_{Ri}, y_{Ri}$) and the robot heading as $\theta_{Ri}$. The robot position with respect to the box is then represented as ($x_b^{Ri}, y_b^{Ri}$) and relative robot heading as $\theta_b^{Ri}$ for each robot \(i = 1, \dots, N\). 
The goal is to transport the box to the goal position $b(T) =$ ($x_G, y_G$) (in the global frame, refer Fig.~\ref{fig:global-local}) in some finite time $T$. 
% The $V$ denotes the decision variables $[p_i (t), v_i(t), \omega_i(t)]$ which incorporates the relative robot positions (w.r.t box) \(p_i(t) = (x_b^{Ri}(t), y_b^{Ri}(t), \theta_b^{Ri}(t) )^\top\), the linear velocity inputs \(v_i(t) \in \mathbf{R}\), and the angular velocity inputs \(\omega_i(t) \in \mathbf{R}\) for each robot \(i = 1, \dots, N\) and for all times \(t \in [0,T]\).
% The optimization is performed over the position trajectories \(p_i(t) = (x_b^{Ri}(t), y_b^{Ri}(t), \theta_b^{Ri}(t) )^\top\), the linear velocity inputs \(v_i(t) \in \mathbf{R}\), and the angular velocity inputs \(\omega_i(t) \in \mathbf{R}\) for each robot \(i = 1, \dots, N\) and for all times \(t \in [0,T]\). 
% \newpage
The corresponding optimization formulation can be given as: 
%\vspace{-3mm} % Adjust this value to your needs

\begin{subequations}
\footnotesize
\begin{align}
\min J(V) \label{eqn:obj} \\
\text{s.t.} \quad
% & x_i(0) = x_{i,0}, \label{eq:boundary_x} \\ 
& b(0) = b_{\mathrm{start}}, \quad b(T) = b_{\mathrm{goal}}, \label{eq:boundary_b} \\ 
& v_{i,\min} \le v_i(t) \le v_{i,\max},  \label{eq:bounds_v} \\
& \omega_{i,\min} \le \omega_i(t) \le \omega_{i,\max}, \label{eq:bounds_w} \\
& \|p_i(t) - p_j(t)\| \le L_b^{(i,j)} \label{eq:caging}
\\
& 
\begin{bmatrix}
\sin \theta_i(t) \\
-\cos \theta_i(t)
\end{bmatrix}^\top
\dot{p}_i(t) = 0, \label{eq:nonholonomic}
\end{align}
\end{subequations}

% \noindent
% for all robots $i=1,\dots,N$ and all times $t\in[0,T]$.
\noindent where $V$ denotes the decision variables $[p_i (t), v_i(t), \omega_i(t)]$ which incorporates the relative robot positions (w.r.t box) \(p_i(t) = (x_b^{Ri}(t), y_b^{Ri}(t), \theta_b^{Ri}(t) )^\top\), the linear velocity inputs \(v_i(t) \in \mathbf{R}\), and the angular velocity inputs \(\omega_i(t) \in \mathbf{R}\) for each robot \(i = 1, \dots, N\) and for all times \(t \in [0,T]\). 

In Eqn.~\ref{eqn:obj}, $J(V)$ represents the objective function which can assume various forms. It can be represented as time of the mission ($T$), energy or control inputs ($v_i, \omega_i$), deviation of box trajectory from the reference trajectory ($b(t)$) or a weighted combination of all of them. 

The constraint given by Eqn.~\ref{eq:boundary_b} refers to the initial and final position of the box. Eqns.~\ref{eq:bounds_v} and \ref{eq:bounds_w} indicate the robot linear and angular speed constraints. The constraint in eqn.~\ref{eq:caging} represents the caging constraint that the distance between two robots placed consecutively should be less than the box length they cover. The non-holonomic constraints of the robot is given by eqn.\ref{eq:nonholonomic}.  

Given the complex contact dynamics it is very difficult to get the analytical form of the objective. Hence, this often results in optimization over simulation, In addition, the number of decision variables quickly scales with the discretization of the time horizon and the number of robots. It is therefore computationally intractable and not scalable. 
Some of the previous works have introduced assumptions, constraints and demonstrated successful transport \cite{tang2024collaborative, ebel2024cooperative}. However, it should be noted that they did not incorporate either the caging constraints or non-holonomic constraints (and the corresponding cost of re-formation) or both. 

Solving the above optimization problem (Eqn.~\ref{eqn:obj}) may also limit the generalizability to the given environmental conditions (terrain, box mass). Therefore, in this paper, we do not optimize over any objectives. Instead, we present a method which uses assumptions (described in Sec.~\ref{sec:intro}), introduces heuristics, and thereby 
achieves the box transport while respecting the constraints. The next section (Sec.~\ref{sec:method}) explains our proposed method (R2P2), and in the subsequent section (Sec.~\ref{sec:num-expts}), we analyze the generalizability of the method against various environment parameters. 

%\vspace{-2mm}
\section{R2P2 Method}\label{sec:method}

Our method (R2P2) decomposes the mission into two primitives: 1) Box rotation and 2) Box translation. The box is first rotated so that the box heading aligns with the vector pointing to the goal-point from the box centroid. 
It is then linearly translated towards that goal-point.
At the start of the mission, we allocate the relative robot position (w.r.t box) to all the robots based on heuristics. 
The heuristics are guided by two factors: 1) Caging condition: The distance between robots on the adjacent edges should be less than the minimum box length. The distance between robots on the same edge should be less than the edge length. 
2) Optimality criterion: The relation between force vector ($||\mathbf{F}_i||$) and the robot position ($\mathbf{r}_i$) is given by the standard force-moment equation (Eq.~\ref{eqn:moment}). Using this equation, the optimal robot placement for box rotation turns out to be the corners. Once the robots start actively pushing the box, the robots near the corners can quickly get out of the box due to uncertain contact dynamics.  Therefore, the robots are assigned locations at some $\delta$ distance from the edges. 

% Using the standard physics equation \ref{eqn:moment} to compute the moment on the center of the box, the optimal contact points are towards the corners of the box in terms of force magnitude required to achieve box rotation. 
% During the course of box transport, anti-clockwise, clockwise and straight motion of the box are required. 
Considering these two criteria, the initial relative robot location assigned for a team of 6 robots is shown in Fig.~\ref{fig:mission-mode}. 
The robots 1, 2, 4, 5 are placed at some $\delta$ distance from the corner, while the robots 3, 6 are placed at the center of the edge. 
Apart from the prevention of box slippage, caging also allows rectification of box motion with minimal adjustments in robot positions \cite{tang2024collaborative}. 

% %\vspace{-6mm}
{\footnotesize
\begin{equation}\label{eqn:moment} 
\mathbf{M} \;=\; \sum_{i=1}^{N} \mathbf{r}_i \times \mathbf{F}_i    
\end{equation}
}

% \begin{equation}\label{eqn:moment}
% \mathbf{M} \;=\; \sum_{i=1}^{N} \mathbf{r}_i \times \mathbf{F}_i    
% \end{equation}
% %\vspace{-3mm}

Once the robots are given an initial relative location, we ensure that the robots stay at that location and face perpendicular to the edge during the entire transport mission. The contact dynamics between the robot and the box lead robots away from their position. Therefore, we check the relative robot position of each robot at a certain interval. If the robot deviates from its respective position, a corrective repositioning maneuver is executed. The robot first rotates so that it is parallel to the edge, then it translates back to its original position. Finally, it reorients itself perpendicular to the nearest box-edge. 

\begin{figure}
    \centering
    \includegraphics[width=0.95\linewidth]{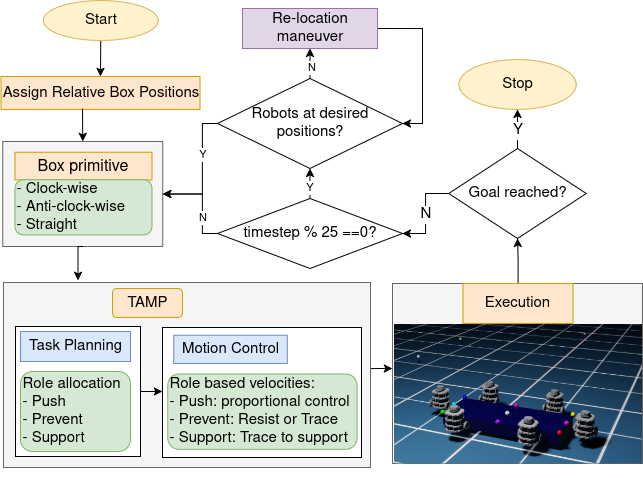}
    \caption{Flowchart of R2P2}
    %\vspace{-7mm} % Adjust this value to your needs
    \label{fig:method-framewk}
\end{figure}

After deciding on the box primitive (rotation or translation), with the assumption of each robot facing perpendicular to the box at the desired position; each robot is allotted a role and subsequently control velocities. We now provide a detailed description of the rules used for deciding box primitive, robot role, and robot velocities. The flowchart of our framework is shown in Fig.~\ref{fig:method-framewk}. 

% \begin{figure}
%     \centering
%     \includegraphics[width=0.95\linewidth]{Figures/framework/coop-transport-Page-1.drawio.png}
%     \caption{Flowchart of R2P2}
%     %\vspace{-7mm} % Adjust this value to your needs
%     \label{fig:method-framewk}
% \end{figure}
%\vspace{-2mm}
\subsection{Box-primitive decision}
% \textbf{First, Mission-Mode decision:}
There are three types of box primitives, \textit{turn-anticlockwise}, \textit{turn-clockwise} and \textit{straight-motion}.  Given the current box center $(x_c,y_c)$, orientation $\theta_c$, and a target
point $(x_t,y_t)$, the required yaw angle is
%\vspace{-2mm}
\[
\theta_R = \operatorname{atan2}(y_t-y_c,\,x_t-x_c).
\]
The box primitive is selected as follows:
\[
\text{Mode} =
\begin{cases}
\texttt{turn-anti-clockwise}, & \theta_R > \theta_c + \epsilon, \\[4pt]
\texttt{turn-clockwise}, & \theta_R < \theta_c - \epsilon, \\[4pt]
\texttt{straight-motion}, & \text{otherwise}.
\end{cases}
\]

\subsection{Role Assignment}
Each robot is categorized into one of the three roles: \{\textit{Push}, \textit{Prevent}, \textit{Support}\}. The rules developed for assigning the roles are based on simple, intuitive principles. The job of \textit{push} robots is to ensure continual box motion towards the goal.
% achieve progress of the mission towards the goal. 
\textit{Prevent} robots resist the motion of the box. Therefore, they prevent excessive increase in speed (linear \& angular) of the box. The support robots do not actively contribute to the box motion. They just ensure that the box does not slip away from the robot formation. 

Let the box length be $l$, box width $w$, global position of the box centroid $\mathbf{c}=(x_b,y_b)$, current box heading $\theta_b$, global position of the robot $i$, $\mathbf{r}_i=(x_{Ri},y_{Ri})$. 
% Along with the length ($l$) and width of the box ($w$), we assume the perfect knowledge of the global position of the box centroid ($\mathbf{c}=(x_b,y_b)$), global position of all the robots ($\mathbf{r}_i=(x_{Ri},y_{Ri})$) and the 
% Assume the global positions of the box center $\mathbf{c}=(x_b,y_b)$ and all robots $\mathbf{r}_i=(x_{Ri},y_{Ri})$ are known.  
% current box heading ($\theta_b$). 
To reason about which robots are aligned within the length sides of the box, we express each robot’s position in the box-aligned coordinate frame. This is obtained by
% %\vspace{-2mm}
\[
\mathbf{p}_i =
\begin{bmatrix}
\ell_i \\ w_i
\end{bmatrix}
= \mathbf{R}(\theta_b)^\top \bigl(\mathbf{r}_i-\mathbf{c}\bigr),
\quad
\mathbf{R}(\theta_b)=
\begin{bmatrix}
\cos\theta_b & -\sin\theta_b \\
\sin\theta_b & \ \cos\theta_b
\end{bmatrix},
\]
% %\vspace{-1mm}
where $\ell_i$ is the coordinate along the box length and $w_i$ is the coordinate along the box width.

Robots with $|w_i| > L/2+\varepsilon$ are designated \emph{Support} by default (they are positioned outside the length sides). For the remaining robots, let $\hat{\mathbf{f}}_i$ denote their pushing direction and $\hat{\mathbf{d}}$ the unit goal direction. 

\begin{itemize}
    \item[\textbf{a)}] \textbf{Box Translation primitive:} 
    Define $s_i = \hat{\mathbf{f}}_i \cdot \hat{\mathbf{d}}$. The role of robot $i$ is
% \[
% s_i = \hat{\mathbf{f}}_i \cdot \hat{\mathbf{d}}.
% \]
    
% %\vspace{-5mm}
\[
\text{Role}_i =
\begin{cases}
\text{Push}, & |w_i|\le L/2+\varepsilon \ \text{and } s_i > 0, \\[4pt]
\text{Prevent}, & |w_i|\le L/2+\varepsilon \ \text{and } s_i < 0, \\[4pt]
\text{Support}, & \text{otherwise}.
\end{cases}
\]

\item[\textbf{b)}] \textbf{Box Rotation primitive:} The torque contribution of robot $i$ about the box center ($\mathbf{c}$) is
% %\vspace{-3mm}

\[
\tau_i = (\mathbf{r}_i-\mathbf{c}) \times \hat{\mathbf{f}}_i
% \quad
% (x,y)\times(u,v) = x v - y u.
\]
% %\vspace{-6mm}

Given the desired torque direction $\tau_{\text{des}} \in \{+1,-1\}$ (anti-clockwise or clockwise), the role of robot $i$ is
% %\vspace{-2mm}
\[
\text{Role}_i =
\begin{cases}
\text{Push}, & \tau_i \,\tau_{\text{des}} > 0, \\[4pt]
\text{Prevent}, & \tau_i \,\tau_{\text{des}} < 0, \\[4pt]
\text{Support}, & \tau_i = 0.
\end{cases}
\]
\end{itemize}

\begin{figure}
    \centering
    \includegraphics[width=0.85\linewidth]{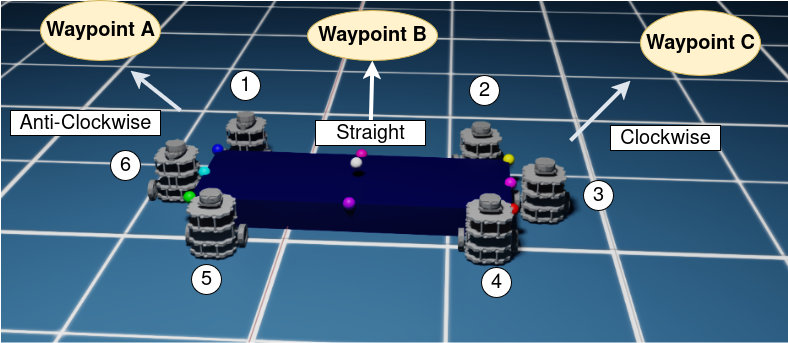}
    \caption{Box primitive for three different target waypoints}
    %\vspace{-6mm} % Adjust this value to your needs

    \label{fig:mission-mode}
\end{figure}

In Fig.~\ref{fig:mission-mode}, the goal waypoint A is to the left of the current box position. Therefore, the box primitive is \texttt{turn-anti-clockwise}. The robots 1 and 4 are assigned the push role, 2 and 5 are the \textit{prevent} robots and the robots 3 and 6 are assigned \textit{support} role. For goal waypoint B, the box primitive is \texttt{straight-motion}. Therefore, the robots 4, 5 are assigned push role, 1 and 2 are the \textit{prevent} robots and the robots 3, 6 are assigned \textit{support} role. 

\subsection{Velocity planning}
The velocity of each robot is computed based on its role and the box primitive. Recall that all the robots face perpendicular to the box and are at their desired relative robot positions (w.r.t box). Since we use differential drive robots, the velocity of robot $i$ is given by linear (forward) speed, denoted as $v_i$ and angular speed (yaw rate), denoted as $\omega_i$.

The robots with push role get positive linear speed, computed with proportional control, where the error term is derived using the current and the required box yaw estimate. 
% The prevent robots need to act if required. Meaning, they should clip undesirable box acceleration (if that occurs), else maintain some distance from the box so that they do not make it difficult for the push robots to do their job. Therefore, 
The \textit{prevent} robots get negative linear and angular speeds based on the following rule. 
% for tracing their relative positions. The 
They trace their relative position when box acceleration is within bounds. We set a lower bound to the speed of the \textit{prevent} robots that prevents undesirable box acceleration. 
% Thus, prevent robots prevent undesirable 
The job of \textit{support} robots is just to support the box from slipping out of the formation. Therefore, they just trace their relative box position. We now express this mathematically.  

Let $\mathbf{x}_c$ and $\mathbf{x}_d$ denote the current and desired box centers, respectively. Let $\theta_c$ denote the current box yaw and $\theta_R$ the required yaw. The translational error ($e_t$) and the yaw error ($e_\theta$) is given by:

%\vspace{-4mm}
{\footnotesize
\[
e_t = \|\mathbf{x}_c - \mathbf{x}_d\|_2.
\]
}
%\vspace{-4mm}
{\footnotesize
\[
e_\theta = \bigl|\,|\theta_c| - |\theta_R|\,\bigr|.
\]
}

% \[
% e_t = \|\mathbf{x}_c - \mathbf{x}_d\|_2.
% \]
% \[
% e_\theta = \bigl|\,|\theta_c| - |\theta_R|\,\bigr|.
% \]
% %\vspace{-7mm}
\begin{itemize}
    \item[\textbf{a)}] \textbf{Push robots:}
The velocity for push robots is given as:
{\footnotesize
\begin{equation}\label{eqn:push-robots}
  (v_p,\, \omega_p)
  =
  \left(
    \max\!\left( \frac{e}{k_p},\, v_{\min} \right),\; 0
  \right),
\end{equation}
}
% %\vspace{-3mm}

% \begin{equation}\label{eqn:push-robots}
%   (v_p,\, \omega_p)
%   =
%   \left(
%     \max\!\left( \frac{e}{k_p},\, v_{\min} \right),\; 0
%   \right),
% \end{equation}
% %\vspace{-4mm}

where $e = e_t$ for box translation and $e=e_\theta$ for box rotation. The gain $k_p$ can be tuned as per the box mass, maximum robot speed, and the friction coefficient of the surface.  $v_{\text{min}} > 0$ ensures continuous motion. 
    
    \item[\textbf{b)}] \textbf{Prevent robots}:
The velocity of the \textit{prevent} robots is given by:
% %\vspace{-2mm}

% \textit{For box translation,} the velocity of the prevent robots is assigned similar to the push robots. 
% \begin{equation} \label{eqn:pr-translate}
%   (v_r,\, \omega_r)
%   =
%   \left(
%     \max\!\left( \frac{e}{k_R},\, v^{R}_{\min} \right),\; 0
%   \right),
% \end{equation}

% Here, $k_R > k_p$ and $v^{R}_{\min} < v_{\min}$ to ensure box motion towards the goal.   

% \textit{For box rotation,} the velocity of the prevent robots is given by:

{\footnotesize
\begin{equation} \label{eqn:pr-rotate}
\begin{bmatrix}
  v_r \\[6pt]
  \omega_r
\end{bmatrix}
=
\begin{bmatrix}
  -\min\!\bigl(\,\|\mathbf{v}_b\| + \omega_b r,\, v^{b}_{\max}\bigr) \\[6pt]
  -\omega_b
\end{bmatrix}.
\end{equation}
}

% \begin{equation} \label{eqn:pr-rotate}
% \begin{bmatrix}
%   v_r \\[6pt]
%   \omega_r
% \end{bmatrix}
% =
% \begin{bmatrix}
%   -\min\!\bigl(\,\|\mathbf{v}_b\| + \omega_b r,\, v^{b}_{\max}\bigr) \\[6pt]
%   -\omega_b
% \end{bmatrix}.
% \end{equation}
% %\vspace{-0.5mm}
where $\mathbf{v_b}, \omega_b$ is the linear and angular speed of the box center. $v^{b}_{\max}$ is set as per the maximum allowable speed of the box.  
    
    \item[\textbf{c)}] \textbf{Support robots:} 
The speed of the \textit{support} robots is given by:
% %\vspace{-6mm}

% \begin{equation} \label{eqn:support-robot}
% \begin{bmatrix}
%   v_s \\[6pt]
%   \omega_s
% \end{bmatrix}
% =
% \begin{bmatrix}
%   -\|\mathbf{v}_b\| + \omega_b r \\[6pt]
%   -\omega_b
% \end{bmatrix}.
% \end{equation}
{\footnotesize
\begin{equation} \label{eqn:support-robot}
\begin{bmatrix}
  v_s \\[6pt]
  \omega_s
\end{bmatrix}
=
\begin{bmatrix}
  -\|\mathbf{v}_b\| + \omega_b r \\[6pt]
  -\omega_b
\end{bmatrix}.
\end{equation}
}

\end{itemize}

% \subsection{Additional Features}
Note that in R2P2, the \textit{support} robots are not a strict requirement, especially for ground based transport. R2P2 is applicable with fewer robots as well, as opposed to caging based transport methods such as \cite{wan2017multirobot, vardharajan2022collective} where caging is a strict requirement for object-transport. 
In the Sec.~\ref{sec:hardware}, we demonstrate that through the hardware experiments, where we use a team of 4 robots to achieve the transport over a flat-terrain. 
%\vspace{-4mm}

\section{Numerical Experiments (in Simulation)} \label{sec:num-expts}
To test the performance of R2P2 on inclined planes as well as on flat terrains, we conduct a set of simulation experiments in a high fidelity virtual environment built using NVIDIA IsaacSim. In all the case studies, we use a team of 6 robots, transporting a box of size 1.35 m x 0.35 m x 0.1 m. 
% \color{red}
The determination of mission success/failure and termination of the mission (simulation or experiment) are given by: \vspace{-0.2cm}
\begin{itemize}\itemsep0em
    \item \textbf{\textit{Success}:} If and when the distance of the box centroid from the goal position is within a certain specified tolerance before the maximum allowed mission time has elapsed, the experiment is terminated; and considered as successful (\textbf{S}). 
    \item \textbf{\textit{Failure-A}:} If the distance of the box centroid from the goal position is not within the specified tolerance when the maximum allowed mission time has elapsed, the mission is terminated; and considered as failure (\textbf{F-A}).
    \item \textbf{\textit{Failure-B}:} If the box slips down the terrain (in the case of uphill/downhill scenarios) or is out of the robot formation (in the case of flat-terrain scenarios) or if there's an inter-robot collision before the maximum allowed mission time has elapsed, the mission is terminated; and considered as failure (\textbf{F-B}).
\end{itemize}

Note that since we are not pursuing time-optimal box transport in this work, a generous maximum mission time (of 25 mins) was specified just as a brute-force termination criteria in the case of robots getting stuck or going into any cyclic motion issues. In practice, none of the experiments in simulation required anywhere close to that amount of time. In the case of missions terminated as failed, it is always observed to occur due to criteria F-B, for our given setup and range of scenarios considered. 

In the remainder of this section, we first explain the baseline used for comparative analysis of R2P2, followed by the description of the case-studies over varying terrains. Then we present the results for the generalizability tests, where we evaluate R2P2 over different friction coefficients and box mass. Subsequently, we present the evaluation results of R2P2 versus baseline across various scenarios. 
Lastly, we provide some discussions on the heuristics in R2P2. 

% \color{black}
% \textcolor{blue}{We terminate the mission based on two termination criteria. The mission is considered successful if the distance of the box centroid from the goal position is within a certain tolerance. And it is considered a failed mission if the box slips down the terrain (in case of uphill/downhill transport) or if the box is out of the robot formation in case of flat-terrain transport.}

% We test the generalizability of R2P2 (proposed method) over various envionment parameters. 

% In this section, we present the numerical experiments conducted to test the generalizability of R2P2. First, we 

% Through a number of simulation experiments, we test R2P2 on all three terrains. The simulations were conducted using NVIDIA IsaacSim simulator. 
% The robots we used were turtlebots, which are commonly available differential drive robots. We also test the generalizability of the method on various friction coefficients and box weights. 
%\vspace{-1mm}
\subsection{Baseline}
As a baseline, we use the Virtual Leader Follower (VLF) method for controlling the robots. The centroid of the box is considered to be the virtual leader, and the team of robots are the followers. The robot formation path is calculated using the reference box path. As a consequence of robot pushing or resisting, the box gets transported. 

\begin{figure}
    \centering
    \includegraphics[width=0.95\linewidth]{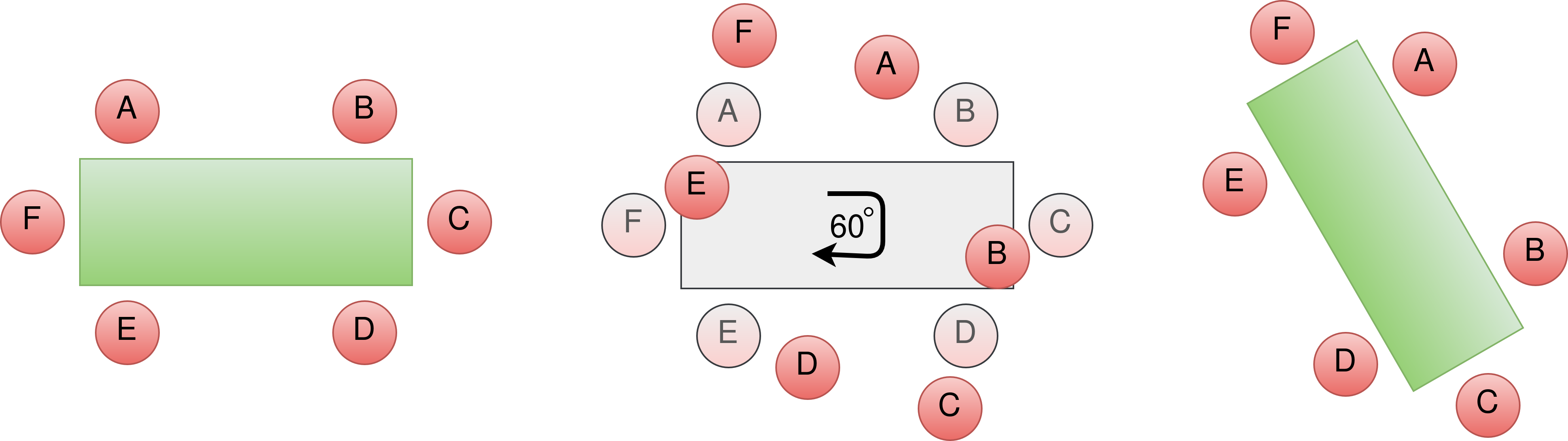}
    \caption{Virtual leader follower. The Figure on the left shows the initial box and robot location. The goal is to rotate the box by $60^{\circ}$. The figure in the middle shows the new robot locations. Through pushing or resisting, the box gets rotated to desired angle as shown in the figure on the right}
    %\vspace{-6mm} % Adjust this value to your needs

    \label{fig:virtual-lf}
\end{figure}

In this paper, we achieve the box transport using box rotation and box translation primitives. 
Accordingly, for box rotation, we compute the desired robots' coordinates corresponding to the rotation of box about the centroid.  
Given the box center ($\mathbf{c}$) and the location for robot $i$ ($\mathbf{r}_i$), the desired robot location ($\mathbf{r}^{'}_i$) is given by Eqn.~\ref{eqn:virtual-lf}. 
Figure \ref{fig:virtual-lf} gives the pictorial representation of VLF. The figure on the left shows the initial position and orientation of the box. To rotate the box anti-clockwise by $60^{\circ}$, the robots plan their new position, highlighted by dark colors in the middle figure. 
Through pushing or resisting, the robots rotate the box by the required angle. The robots get constant linear velocity for box translation.
% For the box translation, the robots get constant linear velocities. 
%\vspace{-3mm}

{\footnotesize
\begin{equation} \label{eqn:virtual-lf}
\mathbf{r}_i' 
= \mathbf{c} 
+ 
\begin{bmatrix}
\cos\theta & -\sin\theta \\
\sin\theta & \cos\theta
\end{bmatrix}
\bigl(\mathbf{r}_i - \mathbf{c}\bigr),
\end{equation}
}

% \begin{equation} \label{eqn:virtual-lf}
% \mathbf{r}_i' 
% = \mathbf{c} 
% + 
% \begin{bmatrix}
% \cos\theta & -\sin\theta \\
% \sin\theta & \cos\theta
% \end{bmatrix}
% \bigl(\mathbf{r}_i - \mathbf{c}\bigr),
% \end{equation}

For fair comparison, we include the re-location mechanism in VLF. After the box primitive changes from rotation to translation, we check the relative box position for each robot. Additionally, during the box translation, we check the robot location at every 25 timesteps, and use our re-location maneuver if the robots deviate from their desired position. 

% \[
% \mathbf{r}_i' 
% = \mathbf{c} 
% + 
% \begin{bmatrix}
% \cos\theta & -\sin\theta \\
% \sin\theta & \cos\theta
% \end{bmatrix}
% \bigl(\mathbf{r}_i - \mathbf{c}\bigr)
% % \quad i=1,\dots,6.
% \]

%\vspace{-3mm}
\subsection{Test over varying terrains}
\begin{figure*}[t]
\begin{minipage}[h]{0.36\linewidth}
    \includegraphics[width=0.85\linewidth]{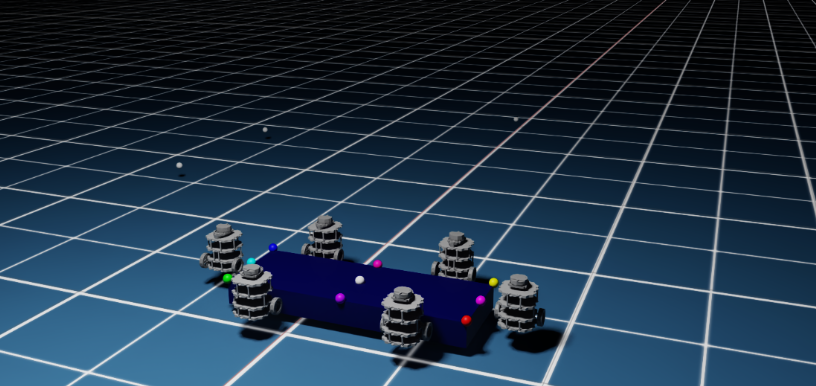}
    \subcaption[]{Flat Terrain Transport}
    \label{subfig:flat-isaac}
\end{minipage}
\begin{minipage}[h]{0.34\linewidth}
    \includegraphics[width=0.85\linewidth]{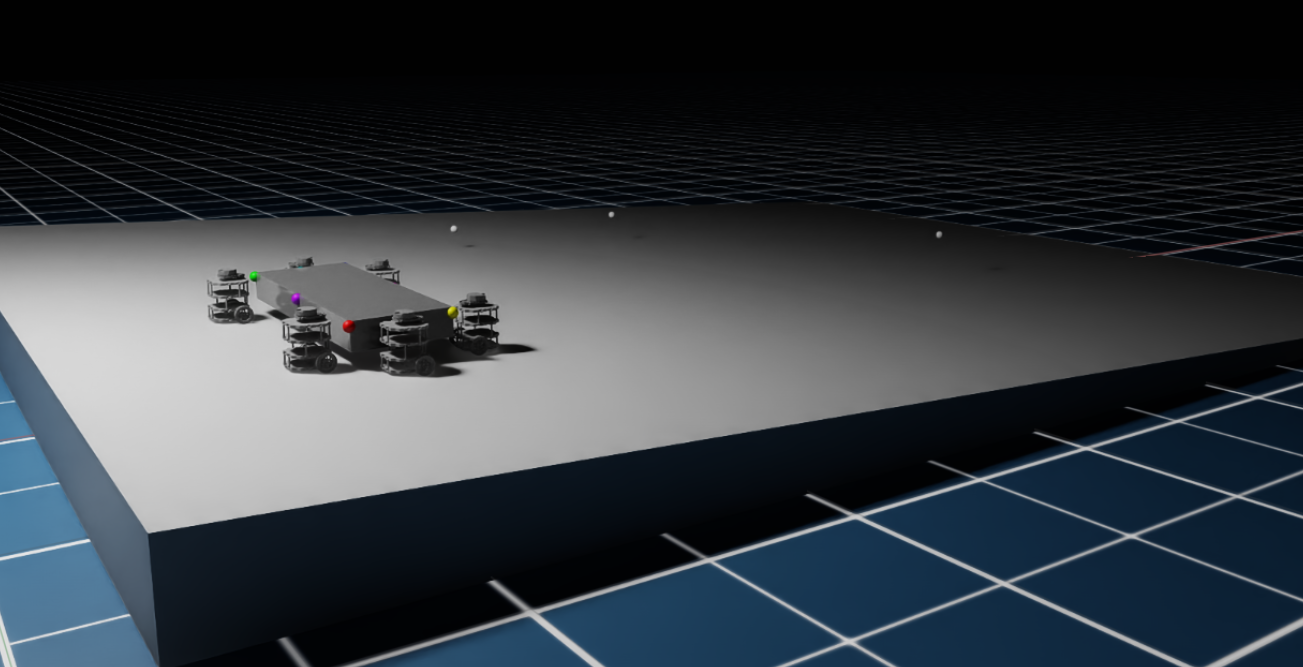}
    \subcaption[]{Down-Hill Transport}
\label{subfig:down-hill-isaac}
\end{minipage}
\begin{minipage}[h]{0.34\linewidth}
    \includegraphics[width=0.85\linewidth]{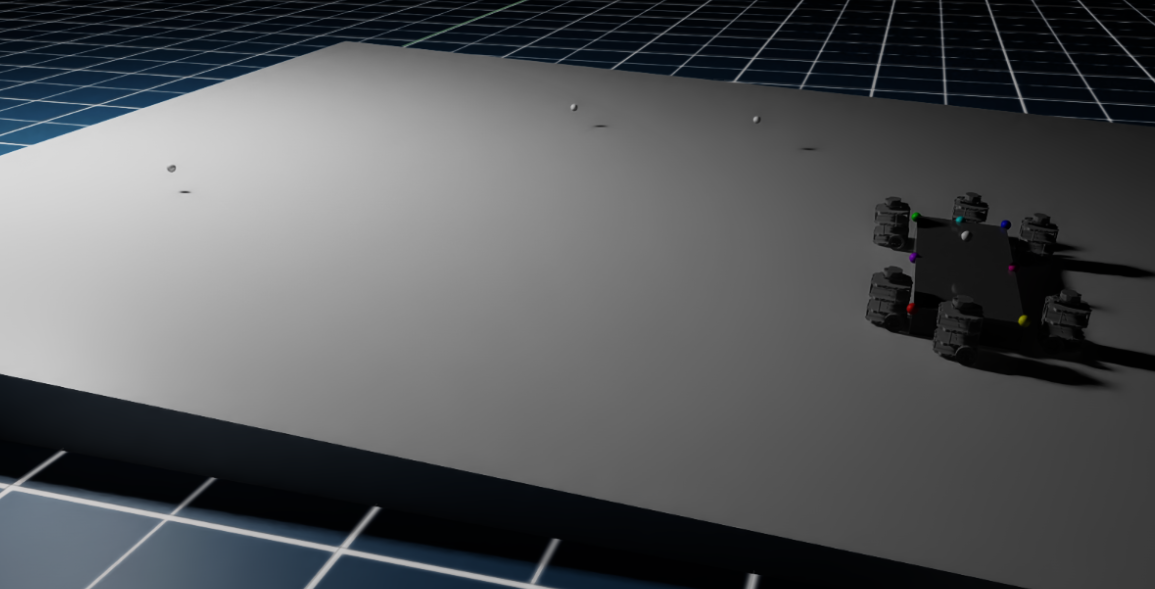}
    \subcaption[]{Up-Hill Transport}
\label{subfig:up-hill-isaac}
\end{minipage}

\begin{minipage}[h]{0.34\linewidth}
    \includegraphics[width=0.85\linewidth]{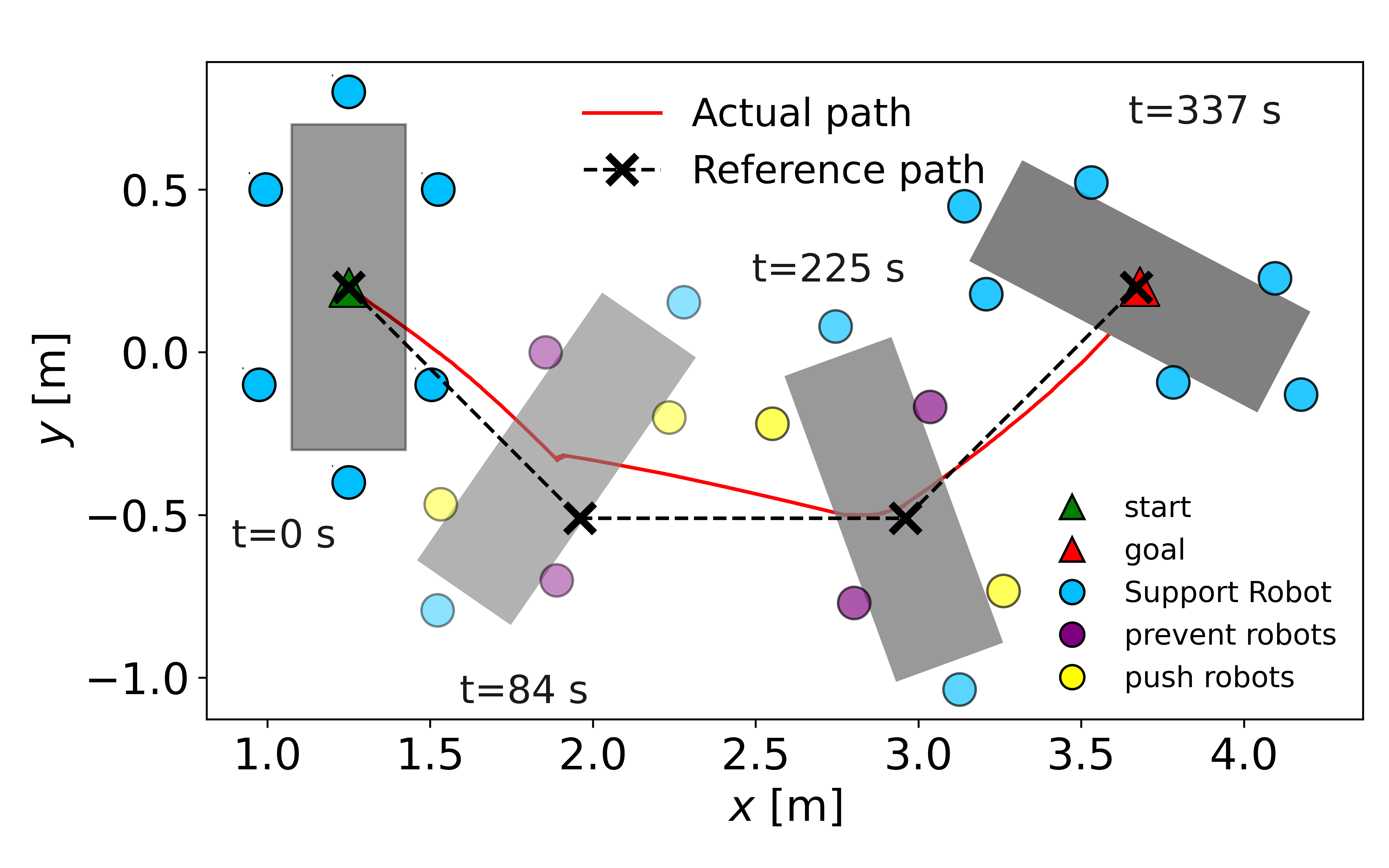}
    \subcaption[]{Flat Terrain Transport}
    \label{subfig:flat-matplotlib}
\end{minipage}
% % \hspace{-5mm}
\begin{minipage}[h]{0.34\linewidth}
    \includegraphics[width=0.95\linewidth]{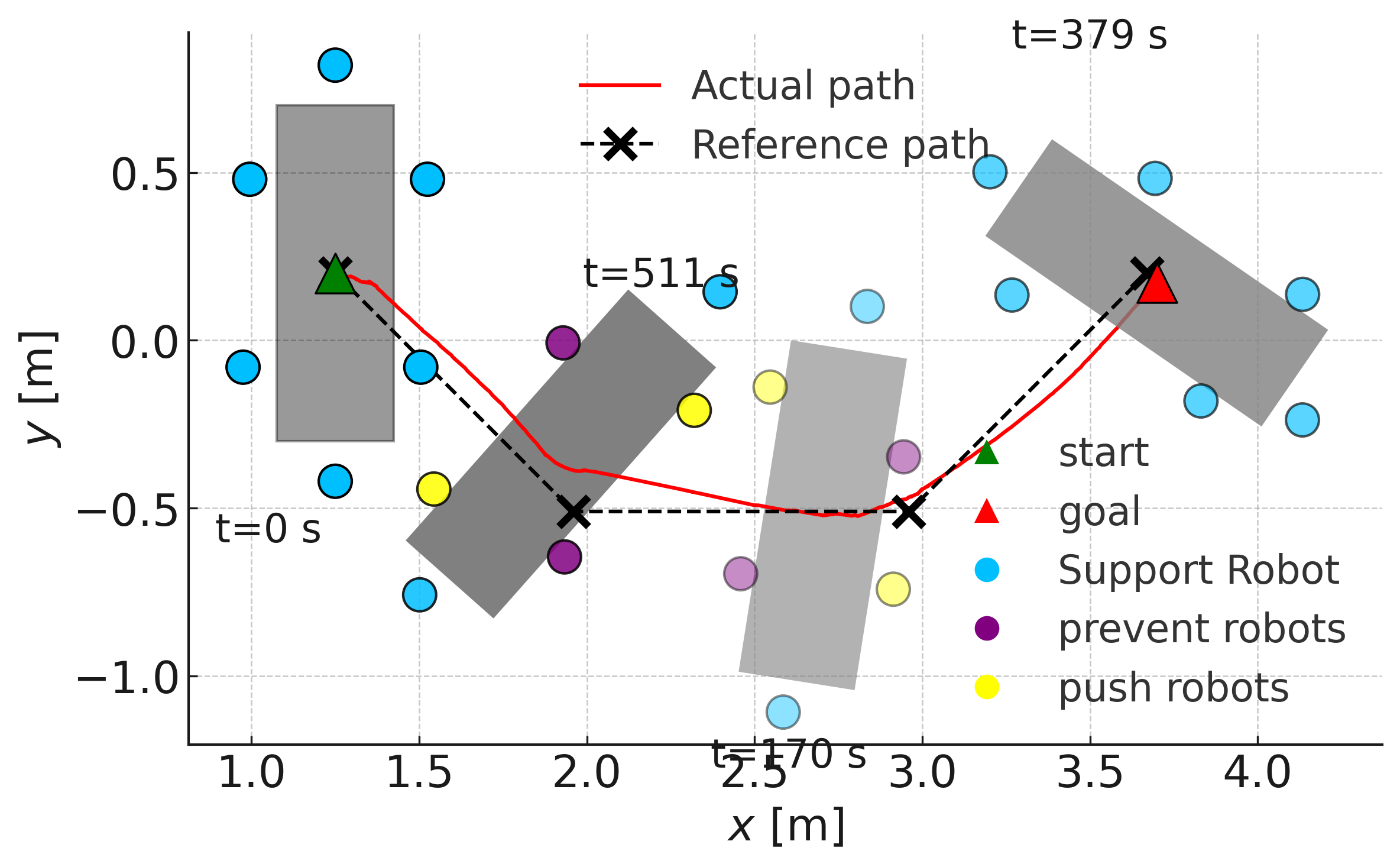}
    \subcaption[]{Down-Hill Transport}
\label{subfig:down-hill-matplotlib}
\end{minipage}
\begin{minipage}[h]{0.34\linewidth}
    \includegraphics[width=0.85\linewidth]{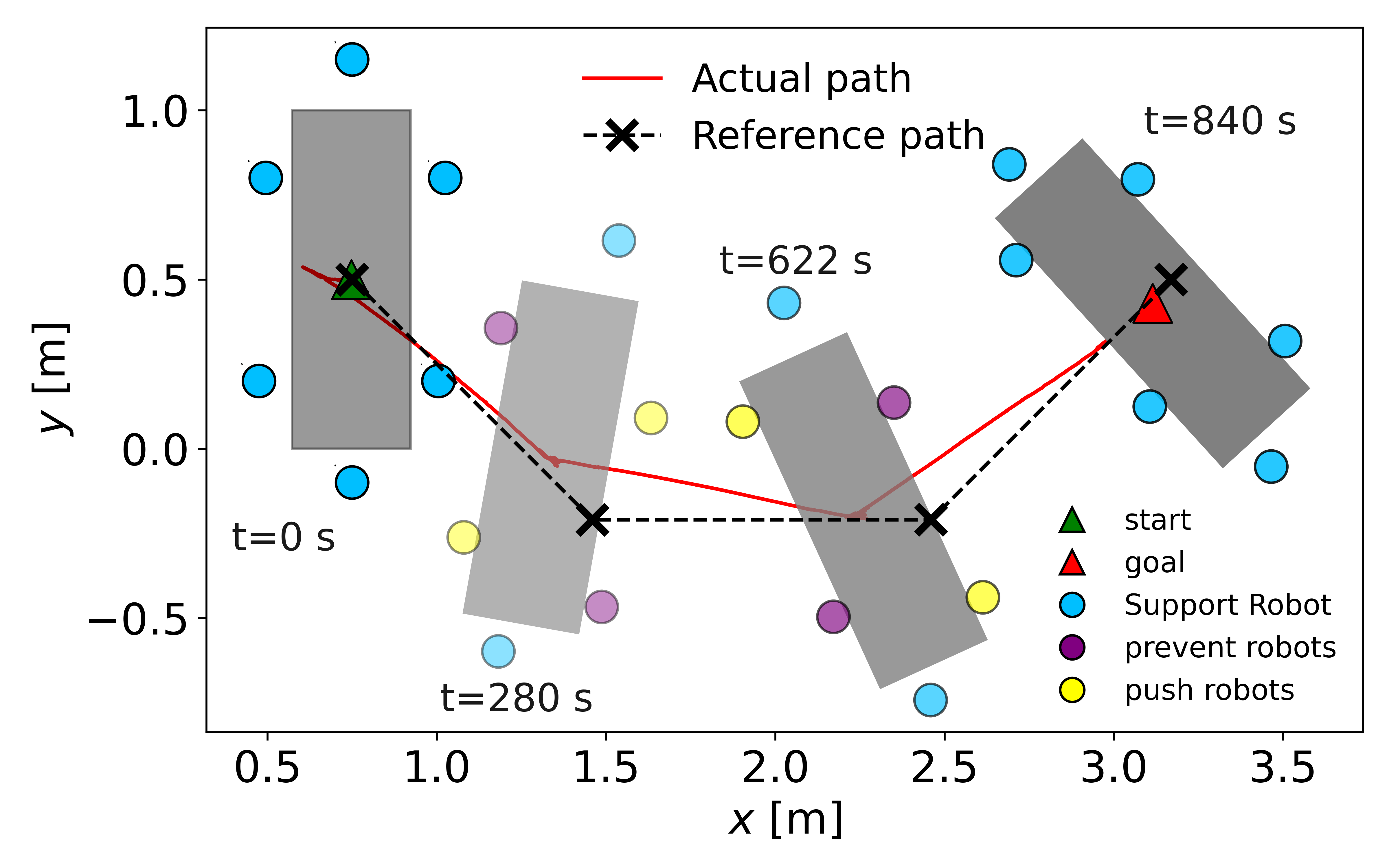}
    \subcaption[]{Up-Hill Transport}
\label{subfig:up-hill-matplotlib}
\end{minipage}

    \caption{Demonstration of R2P2 over various terrains in IsaacSim. The up-hill and down-hill transport is over an inclined plane of 5$^{\circ}$}
    %\vspace{-3mm} Inclined planes and Flat Terrain in IsaacSim % Adjust this value to your needs
    \label{fig:incl-flat}
\end{figure*}

We first test R2P2 for box transport over flat horizontal surface and two inclined surfaces, uphill and downhill planes. A common reference path is designed for all the case studies (shown in Fig.~\ref{subfig:flat-matplotlib}). Starting from an initial position, we provide the intermediate waypoints to the goal location. 
The box mass is set to 2 kg, the friction coefficient of all the three surfaces with respect to the box are set to 0.01. The slope of the inclined planes are 5 degrees. These conditions are sufficient to test the caging over the inclined planes. In the absence of proper caging, the box slips down the terrain.
Figures \ref{subfig:flat-isaac}, \ref{subfig:down-hill-isaac}, \ref{subfig:up-hill-isaac} respectively show the environment setup in IsaacSim for the three scenarios.

A successful transport mission is accomplished for all three scenarios by the team of 6 robots. 
Note that the same heuristic parameters (gain $k_p$) have been used in all the scenarios. 
We came-up with the parameters through offline trial and error in our preliminary experiments, and they have not been tuned for each individual scenario reported in the paper.  
% We did not tune the parameters for individual scenarios. 
% Note that the heuristic parameters were not tuned for a particular scenario. 
The resultant box trajectories for each scenario, along with the reference path is shown in the figures on the bottom (Fig.~\ref{subfig:flat-matplotlib}, \ref{subfig:down-hill-matplotlib}, \ref{subfig:up-hill-matplotlib}). The physical time elapsed (simulated) at various instances of the mission is also reported in the figures. The overall time required for flat-terrain transport, down-hill and up-hill transport is found to be 337 s, 385 s, 840 s, respectively. \
The down-hill transport time is marginally higher than flat-terrain, while the up-hill transport mission takes more time. 
Without a time-optimal tuning of heuristics, this is expected since the robots are working against gravity in case of up-hill transport.  
% Even with the same control cost, the up-hill movement of robots by itself takes a lot of time. 

% textcolor{red}{The downhill and uphill transport missions take more time than the flat terrain because the low friction coefficient introduces a lot of uncertainty, which leads to increased number of relocation maneuvers.} 
In addition, the plots also show the robot roles during the mission. As can be seen in the plots, the roles of the robots change during the mission as per the required box primitive. 
% The plots also show the robot role for each robot. During clockwise
% Further in this section, we will analyze the relation between the heuristic parameters, the number of relocation maneuvers and the simulation time. 
The plots shown in Fig.~\ref{subfig:flat-matplotlib}, \ref{subfig:down-hill-matplotlib}, \ref{subfig:up-hill-matplotlib} indicate that the resultant box trajectory closely aligns with the reference box trajectory in all three scenarios.

%\vspace{-1.5mm}
\subsection{Test for generalizability}
In this second case study, we evaluate the generalizability performance of R2P2 across different friction coefficients and the box mass. 
% Specifically, we test R2P2 for up-hill and down-hill scenarios. 
For each scenario (flat terrain, up-hill, down-hill), we test R2P2 for a range of friction coefficient values and box mass. 
The dynamic friction coefficient ($\mu$) takes values from $\{0.001, 0.01, 0.1\}$. $\mu = 0.001$ makes the terrain very slippery, while $\mu=0.1$ makes it rough. The static friction is fixed to $0.9$.
The box mass takes values from $\{0.5, 2, 6\}$ kg. 
Given the range of surface inclination and surface friction tested, the maximum mass of $6$ kg is used based on observed robot capacity in preliminary experiments. Note that for flat terrain with friction coefficient of 0.1, R2P2 is observed to be able to transport upto 15 kg. 
% and the box mass takes values  from $\{0.5, 2, 6\}$ kg. 
% Based on numerical experiments, we observed that the maximum box mass that could be carried by the robot team over all the scenarios was 7 kg. 
The goal location is positioned at a bearing of $45^{\circ}$ and at a range (distance) of $1$ m with respect to the initial box location. 

% Overall, R2P2 achieved successful transport in all the cases. Successful transport means that the robot team was able to transport the box to its goal location in some finite time. It also implies that the robots were able to maintain their relative position, thereby ensuring consistent mission progress. The caging constraints were also adhered to during up-hill, down-hill scenarios which prevented box from slipping down the terrain. We now report the mission time for all the cases in each scenario. 

Overall, R2P2 is observed to achieve successful transport in all the cases. Along with being able to transport the box to its goal location in reasonable time frame, the robots are also observed to maintain their relative position, thereby ensuring consistent mission progress. The caging constraints are also adhered to during up-hill, down-hill scenarios which prevented box from slipping down the terrain. The mission times for all the case studies over all the scenarios are given in Table \ref{tab:terrain_combined}. In general, we observe that the mission time increased with increase in box mass for the flat and up-hill surfaces. No specific trend is observed for down-hill scenarios. 
\begin{table}[h]
\centering
{\footnotesize
\begin{tabular}{
    >{\raggedright\arraybackslash}p{1.7cm}
    >{\centering\arraybackslash}p{1.2cm}
    >{\centering\arraybackslash}p{0.7cm} 
    >{\centering\arraybackslash}p{0.7cm} 
    >{\centering\arraybackslash}p{0.7cm}
}
\toprule
\textbf{Terrain Type} & \textbf{Mass (kg)} 
& $\mu = 0.001$ & $\mu = 0.01$ & $\mu = 0.1$ \\
\cmidrule(lr){3-5}
 &  & \multicolumn{3}{c}{\textbf{Mission Time (s)}} \\
\midrule
\multirow{3}{*}{Flat} 
 & 0.5 & 77.7  & 76.02 & 69.91 \\
 & 2.0 & 85.34 & 82.89 & 93.64 \\
 & 6.0 & 116.49 & 125.58 & 179.2 \\
\midrule
\multirow{3}{*}{Down-hill} 
 & 0.5 & 78.49 & 62.35 & 72.13 \\
 & 2.0 & 72.34 & 77.94 & 71.53 \\
 & 6.0 & 79.97 & 79.53 & 74.33 \\
\midrule
\multirow{3}{*}{Up-hill} 
 & 0.5 & 244.38 & 232.81 & 267.83 \\
 & 2.0 & 266.44 & 233.26 & 268.18 \\
 & 6.0 & 292.96 & 299.59 & 323.27 \\
\bottomrule
\end{tabular}
}
%\vspace{-2mm}
\caption{Generalizability over terrain type, surface friction coefficient ($\mu$) and object mass reported in terms of \textit{mission time} (s); the up-hill down-hill terrains have an incline of 5$^{\circ}$}
\label{tab:terrain_combined}
%\vspace{-5mm} Generalizability across the surface friction, terrain type
\end{table}

\subsection{Comparison with Baselines}
% The focus of this paper was to develop a new TAMP approach to achieve box-transport using non-holonomic robots on varying terrains. 
% We do not optimize over any objectives such as mission time, control cost or deviation from reference path. 
% The objective was to develop a generalized method for box-transport using non-holonomic robots.
% Therefore, the metric of comparison is mission success or failure. 
Here we compare R2P2 to the baseline VLF method. Performance is compared in terms of success or failure as defined at the start of Section \ref{sec:num-expts}. This is because in the current settings, we do not optimize either of the methods in terms of any other metrics such as time or energy. Since we use box rotation and translation primitives, we design the case-studies for the comparison based on the bearing angle of the goal-point with respect to the initial box heading. The applicability of the method for any potential trajectories is contingent on the success or failure of the method for all required box rotation angles. 
% Once the box heading aligns with the axis connecting the CoM with the waypoint, box translation can take over. 
%
A box of mass 0.05 is used in this case study. We consider 4 variations in the bearings. Case-1, bearing: $15^{\circ}$; Case-2, bearing: $30^{\circ}$; Case-3, bearing: $45^{\circ}$; Case-4 bearing: $60^{\circ}$. We test each case on all three scenarios, namely flat-terrain, up-hill and down-hill. The static and dynamic friction coefficient in each case are set to 0.9 and 0.1 respectively.

\begin{table}[h!]
\centering
\footnotesize

\begin{subtable}{\linewidth}
\centering
\begin{tabular}{c|cc|cc|cc}
% ... first table content ...
\toprule
\begin{tabular}{@{}c@{}} \textbf{Case No.} \\ \textbf{(bearing)} \end{tabular}
& \multicolumn{2}{c|}{\textbf{Up-hill}} 
& \multicolumn{2}{c|}{\textbf{Down-hill}} 
& \multicolumn{2}{c}{\textbf{Flat-terrain}} \\
\cmidrule(lr){2-3} \cmidrule(lr){4-5} \cmidrule(lr){6-7}
& VLF & R2P2 & VLF & R2P2 & VLF & R2P2 \\
\midrule
1 ($15^{\circ}$)& S & S & S & S & S & S \\
2 ($30^{\circ}$)& S & S & S & S & S & S \\
3 ($45^{\circ}$)& F & S & F & S & F & S \\
4 ($60^{\circ}$)& F & S & F & S & F & S \\
\bottomrule

\end{tabular}
\caption{Comparison with baseline: mission success defined by being within a threshold proximity of 20 cm to the goal position}
\label{tab:vlf_proposed}
\end{subtable}

\bigskip

\begin{subtable}{\linewidth}
\centering
\begin{tabular}{c|c|c|c}
% ... second table content ...
\toprule
\begin{tabular}{@{}c@{}} \textbf{Case No.} \\ \textbf{(bearing)} \end{tabular}
& \textbf{Up-hill} & \textbf{Down-hill} & \textbf{Flat-terrain} \\
\midrule
1 ($15^{\circ}$) & S & S & S \\
2 ($30^{\circ}$) & S & S & S \\
3 ($45^{\circ}$) & S & S & S \\
4 ($60^{\circ}$) & S & S & S \\
5 ($75^{\circ}$) & S & S & S \\
\bottomrule

\end{tabular}
\caption{Evaluation of R2P2 with tighter tolerance: mission success defined by being within a threshold proximity of 10 cm to the goal position}
\label{tab:tight_tol}
\end{subtable}

\caption{Results under different terrains and bearing angles. The friction coefficients are set to 0.9 and 0.1 for static and dynamic coefficients respectively. The bearing of the goal position w.r.t the initial position for each case is indicated with brackets in the first column. (S = Success, F = Failure). }
% The mission is considered a \textit{success} if the location of box centroid is within the indicated threshold distance to the goal-position.}
\label{tab:all_results}
\end{table}

The results, in terms of success or failure are reported in Table \ref{tab:vlf_proposed}. As mentioned before, the mission is considered success if the distance between the box centroid and goal-position is within a threshold. We set the threshold to 20 cm for the results in Table~\ref{tab:vlf_proposed}
% \textcolor{red}{Define success/failure here}. 
The Table shows that VLF method (baseline) is successful only for the first two cases over all the three surface inclination scenarios. In contrast, R2P2 is observed to be successful in all the cases over all the three surface inclination  scenarios. 
VLF is found to demonstrate the second mode of Failure (\textbf{F-B}). The robot formation loses the box, which cause the box to slip out and down in the cases of up-hill and down-hill transport, and/or result in an inter-robot collision.
We evaluate our method on the same case-studies but with tighter tolerance of 10 cm and the results are shown in Table.~\ref{tab:tight_tol}. R2P2 achieves successful transport over all the cases even with the tighter goal reaching tolerance of 10 cm, as seen from Table \ref{tab:tight_tol}. 
% In the above case-studies, the tolerance of the 

% VLF is similar to open-loop co
VLF method is similar to open-loop control.
Each robot gets its goal position 
and the outcome of robot navigation achieves the box transport. 
% and the box transport is a result of passive or active pushing. 
It has shown success when used with omni-directional robots \cite{wan2017multirobot}, where the magnitude of the force applied is not dependent on the relative robot-heading (w.r.t the box edge). In case of non-holonomic robots, the magnitude of force applied depends on the current heading of the robot. The contact force is maximum when the robots are perpendicular to the edge, and the force decreases with other relative heading angles.  
% This results in unequal force being applied to the box from different positions. 
Since VLF does not explicitly account for the current robot heading in the motion control, the robots end-up applying forces of different magnitudes from their respective positions. This causes conflicting box movements, which leads the open loop control in VLF to causing robots lose track of their relative positions; this in turn results in either inter-robot collisions or violation of caging constraint (box slips out) and a failed mission. 

We also analyzed the transportation capacity of both methods in terms of box mass. We observed that on the flat-terrain, a team of $6$ robots with R2P2 can transport a box of mass upto $15$ kg, while with VLF the team is able to transport a box of mass only upto $0.05$ kg. This inferior capability of VLF is attributed to robot heading being not aligned perpendicular to the box.

\subsection{Discussion: Robot state, control analysis during box rotation.}

R2P2 is a decentralized role-based, interpretable MR-TAMP formulation which does not use black-box learning or optimization. Here, we provide the interpretation or analysis of the evolution of the robot controls in the box-transport mission.
% Here, we provide the interpretation of the robot controls during a box-rotation maneuver. 
R2P2 achieves box transport by box rotation and box translation. Box translation is relatively straightforward in comparison to box-rotation. Therefore, in this section we analyze the evolution of robot state and controls during a box rotation over a flat-terrain. 
Recall that each robot is assigned either a \textit{push}, \textit{prevent} or \textit{support} role. The \textit{push} and \textit{prevent} robots actively engage in box motion, while the \textit{support} robots do not (Sec.~\ref{sec:method}). Therefore, the speed and acceleration of box is primarily dependent on the speed of \textit{push} and \textit{prevent} robots. 

\begin{figure}
    \centering
    \includegraphics[width=0.95\linewidth]{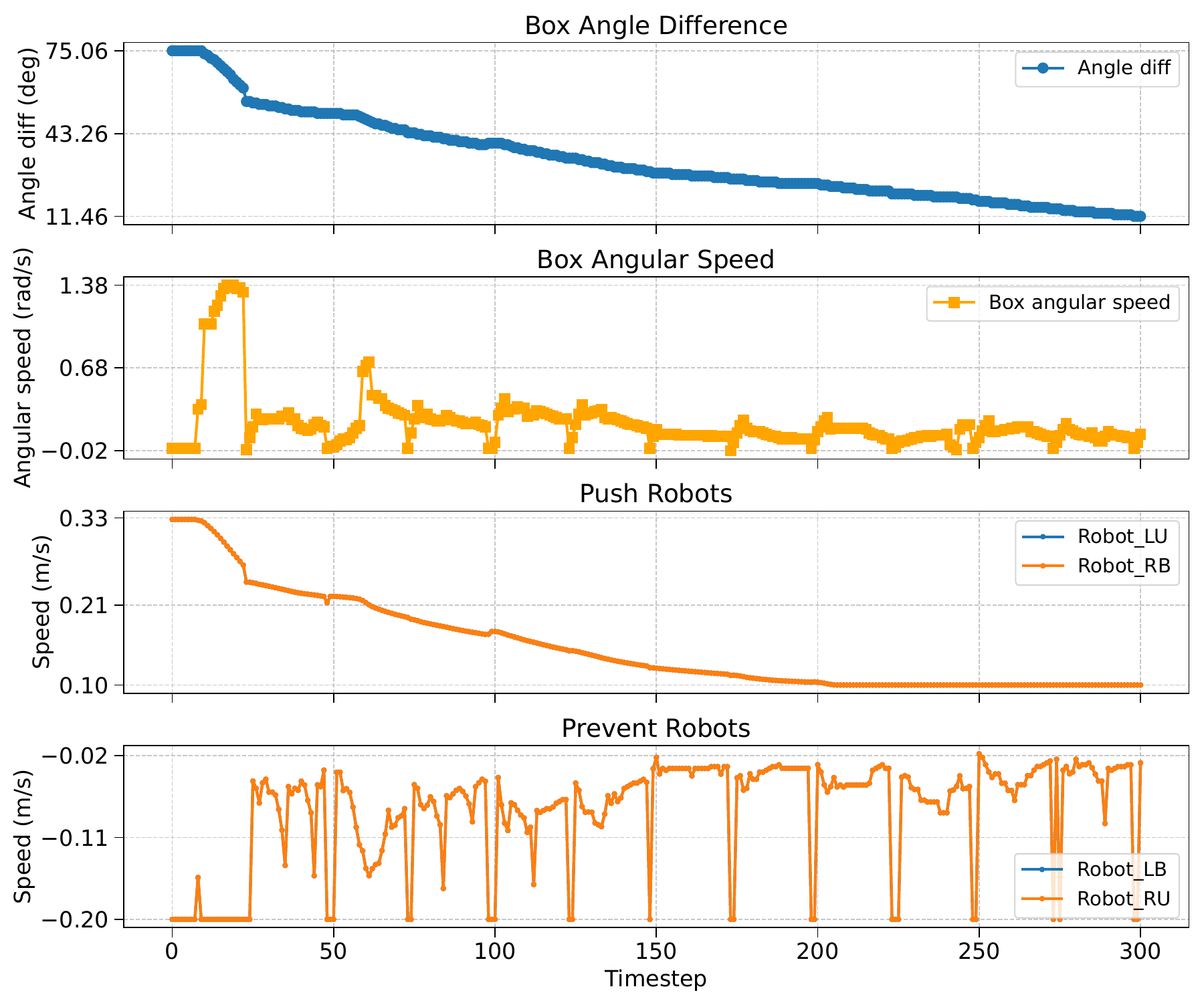}
    \caption{Analysis of the robot states and controls, categorised by their role over the box-rotation of 75$^{\circ}$. The mission is over flat-terrain with dynamic and static friction coefficients being 0.9 and 0.001 respectively.}
    \label{fig:track-speeds}
\end{figure}

% R2P2 gives a new formulation with lesser decision variables and robust with respect to caging constraints. sss

Figure \ref{fig:track-speeds} shows the speed of the \textit{push} and \textit{prevent} robots over the timesteps when rotating a box by $75^{\circ}$. The first plot on the top shows the error (in degrees) between the current box heading and the required box heading angle over the timesteps. 
Initially the error is around 75 $^\circ$, and the error drops down with increasing timesteps. The angular tolerance for the rotation is 11 $^\circ$ (0.2 rad) and the first plot shows that the angular-difference curve starts to flatten out at around 11.46 $^\circ$.
The second plot shows the box angular speed. The angular speed shows a peak at around 25$^\text{th}$ timestep. This peak is attributed to the initial transition of the box from stationary state. Over the mission timesteps, the box angular speed decreases in general, which is expected given that the motion control of \textit{push} robots is based on proportional control. 
As the angular error (first plot) decreases, the speed of the \textit{push} robots decrease (third plot). As a result, the push force decreases over timesteps causing the box angular speed to decrease in general. The second plot also shows occasional troughs. They correspond to the re-positioning maneuver, when the robots get back to their desired relative-position. 
The marginal decrease in the angular speed of the box (second plot) can also be explained with the fourth plot which shows the linear speed of the \textit{prevent} robots. When the speed of the \textit{prevent} robot is more negative, they are moving back faster thereby allowing smooth box rotation. As the speed of the \textit{prevent} robot goes towards zero, they start obstructing the box motion. The fourth plot shows a marginal increase in the speed of the \textit{prevent} robots over the timesteps, similar to the marginal decrease in box-angular speed as observed in the second plot. The troughs in the second plot also align with the fourth plot. It explains the acceleration observed in the box angular motion as a result of the faster backward motion of \textit{prevent} robot.

\section{Sensitivity Analysis}
 
To understand the performance impact of the main heuristic parameters in R2P2 subject to variations in the environment properties (box, target, and surface properties), we conduct a sensitivity analysis. This is performed using the Taguchi L9 crossed array experiment. We develop and use pybullet implementation of R2P2 to conduct the sensitivity analysis. For the purposes of this analysis, we simplify the relocation implementation. Instead of computing collision free path to the desired relative position and orientation, we teleport the robots using pybullet APIs. We penalize each repositioning by adding a fixed penalty time.  
The heuristic inner array (in the Taguchi L9 experiments) comprises four control factors:
rotation gain~$k_{\text{rot}} \rightarrow 1/k_p$ (Eq.~\ref{eqn:push-robots}), translation gain~$k_{\text{trans}} \rightarrow 1/k_p$ (Eq.~\ref{eqn:push-robots}),
minimum velocity~$v_{\text{min}}$ (Eq.~\ref{eqn:push-robots}), and contact placement~$c$,
with maximum velocity $v_{\text{max}} = 0.22$~m/s. The contact placement ($c$) refers to the distance of the robots from the side edge. 
The environment outer array comprised surface friction~$\mu$,
box mass~$m$, target orientation~$\theta$, and target
distance~$d$.
The full factor ranges and levels are given in Table~\ref{tab:taguchi}.
The crossed L9$\times$L9 design yields 81 deterministic simulation
runs per terrain condition.
Task completion time~$y$ (seconds) is used as the performance metric,
with lower values indicating faster transport.
Main effects were computed on successful runs only; failed runs
were penalised with a ceiling value of $y = 600$~s in the
signal-to-noise calculation but excluded from the mean effect plots.

\begin{table}[ht]
\centering
\caption{Taguchi L9 crossed array design parameters and levels ($9 \times 9 = 81$ runs).}
\label{tab:taguchi}
\begin{tabular}{llccc}
\toprule
\textbf{Type} & \textbf{Parameter} & \textbf{L1} & \textbf{L2} & \textbf{L3} \\
\midrule
\multirow{4}{*}{Control}
  & Rotation gain $k_{\text{rot}}$      & 0.125 & 0.250 & 0.500 \\
  & Translation gain $k_{\text{trans}}$ & 0.200 & 0.300 & 0.400 \\
  & Min.\ velocity $v_{\text{min}}$ (m/s) & 0.100 & 0.150 & 0.200 \\
  & Contact placement $c$ (m)           & 0.050 & 0.175 & 0.300 \\
\midrule
\multirow{4}{*}{Noise}
  & Friction $\mu$               & 0.4 & 0.7 & 0.9 \\
  & Mass $m$ (kg)                & 0.1 & 0.3 & 0.5 \\
  & Target angle $\theta$ (deg)  & 5   & 10  & 15  \\
  & Target distance $d$ (m)      & 0.2 & 0.4 & 0.6 \\
\bottomrule
\end{tabular}
\end{table}

\begin{figure}[ht]
    \centering
    \includegraphics[width=\linewidth]{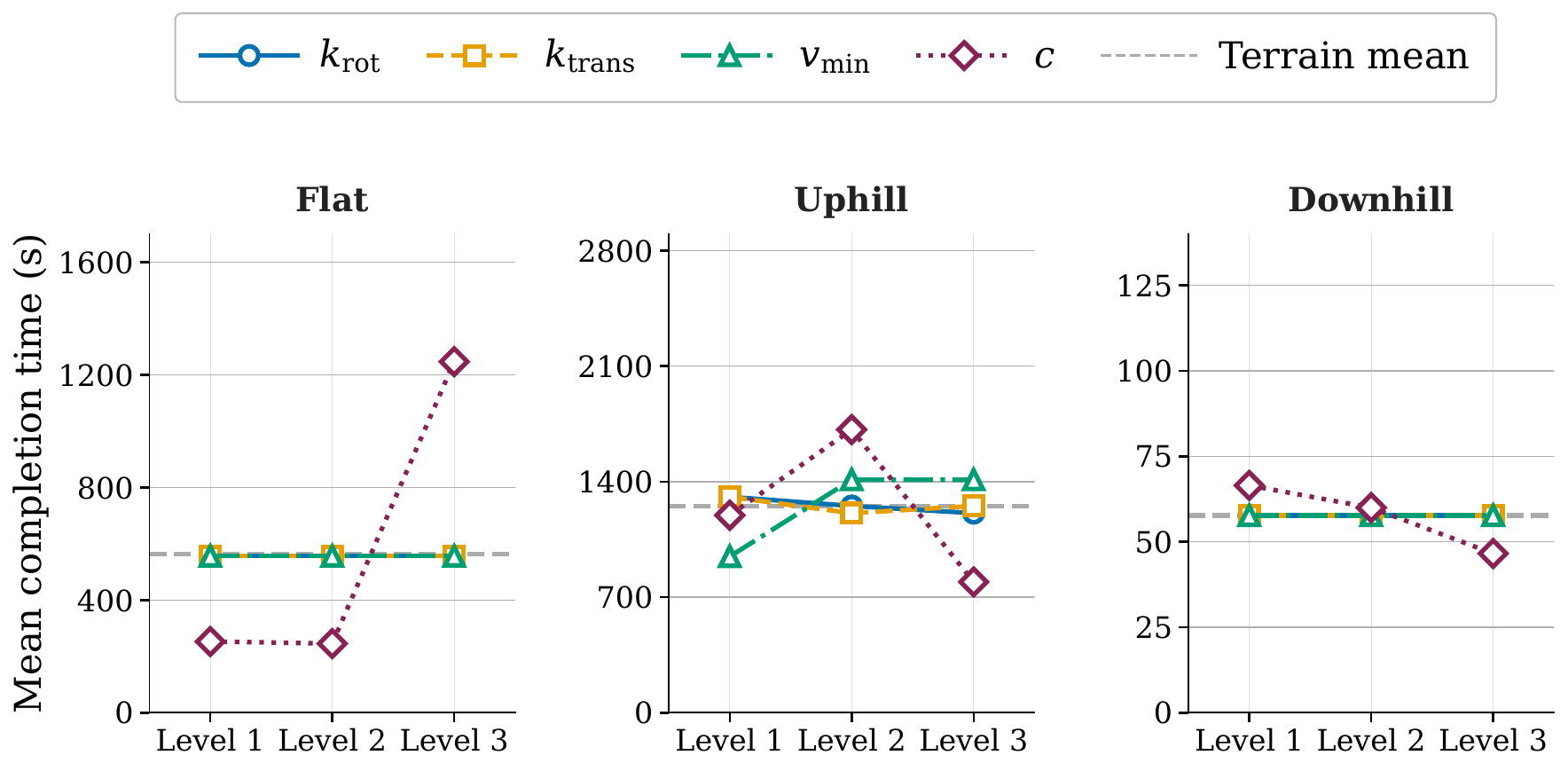}
    \caption{Main effects of heuristic control parameters on task
    completion time across flat, uphill, and downhill terrain.
    Each line corresponds to one factor (see legend); the x-axis
    represents factor level (see Table~\ref{tab:taguchi} for
    level values).
    The dashed line marks the terrain-wide mean.
    % $\Delta$ values are annotated at Level~3 where non-zero.
    % Uphill results are computed on 66 of 81 successful runs.
    }
    \label{fig:heuristic}
\end{figure}

\begin{figure}[ht]
    \centering
    \includegraphics[width=\linewidth]{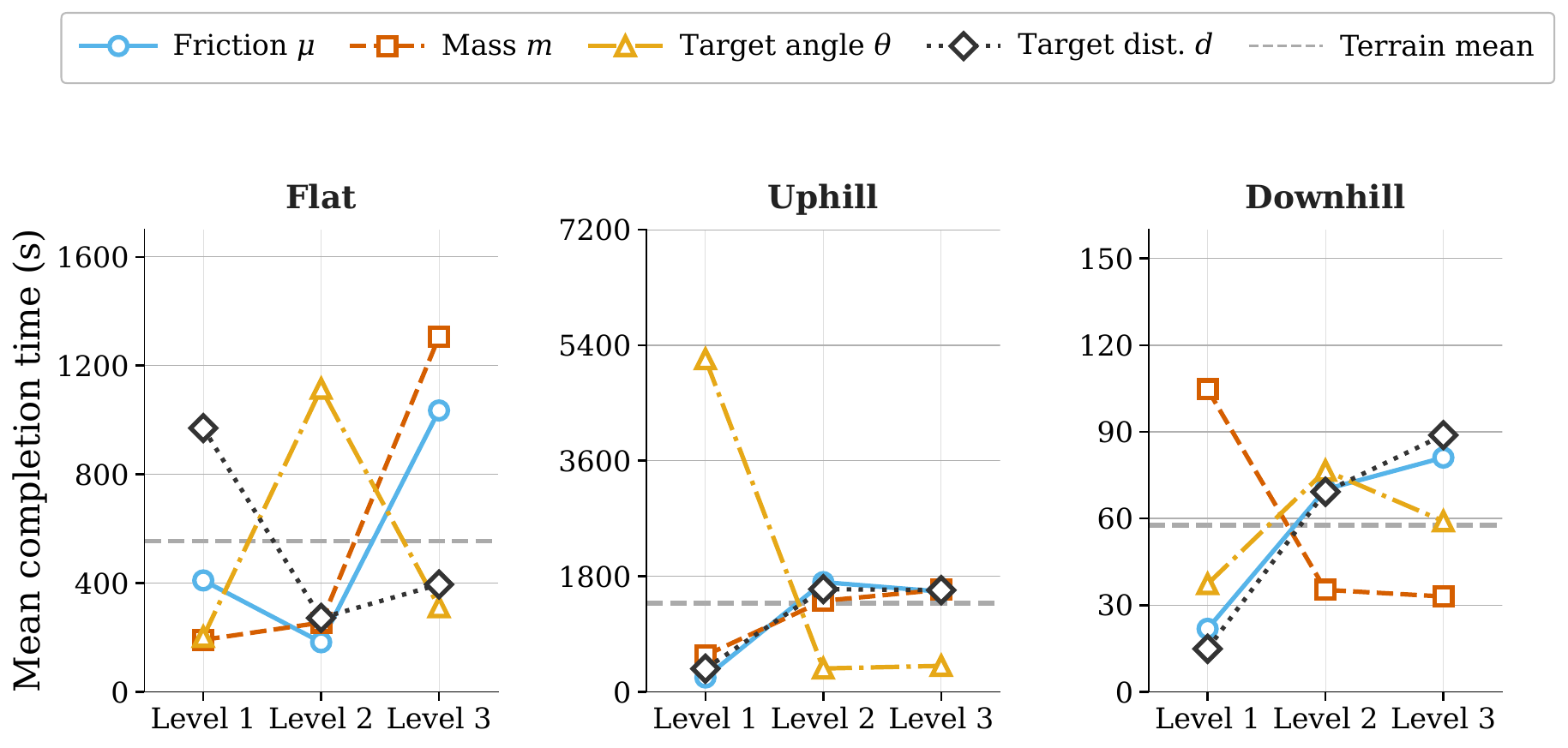}
    \caption{Main effects of environment factors on task completion
    time across flat, uphill, and downhill terrain.
    Layout and notation as in Fig.~\ref{fig:heuristic}.}
    \label{fig:env}
\end{figure}

Figures~\ref{fig:heuristic} and~\ref{fig:env} present the
heuristic and environment main effects respectively across
all three terrain conditions.
Each panel shows four color-coded factor profiles plotted against
factor level (1 -- 3); the dashed horizontal line marks the
terrain-wide mean completion time.
The effect range~$\Delta$ -- the difference between the maximum
and minimum level means -- quantifies each factor's influence
and is annotated at Level~3 where non-zero.
 
% ── Figure 1 ──────────────────────────────────────────────
 
% ── Figure 2 ──────────────────────────────────────────────
 
% ── Flat terrain ──────────────────────────────────────────
\subsection{Flat Terrain}

All 78 of 81 runs completed successfully on flat terrain
(three failures all share $c = 0.300$~m).
The overall mean completion time among successful runs is 556.1~s.

Among the heuristic parameters (Fig.~\ref{fig:heuristic}, left),
$k_{\text{rot}}$, $k_{\text{trans}}$, and $v_{\text{min}}$
each produce $\Delta = 0$~s across all tested levels,
confirming that within the tested ranges these parameters have
no measurable influence on task completion time.
Contact placement~$c$ is the sole heuristic factor of
consequence ($\Delta = 1002$~s).
Performance at $c = 0.050$~m and $c = 0.175$~m is comparable
(253~s and 246~s respectively), while $c = 0.300$~m yields
1247~s among runs that completed and causes three outright
failures, identifying a hard stability boundary near the outer
edge of the contact range.

Among the environment factors (Fig.~\ref{fig:env}, left),
all four exhibit non-zero and non-monotonic profiles.
Mass~$m$ is dominant ($\Delta = 1115$~s, 31.1\% of total
environmental spread), with the heaviest box requiring on
average 6.8$\times$ longer than the lightest.
Target angle~$\theta$ produces the second-largest range
($\Delta = 913$~s, 25.5\%), with the intermediate level of
10\textdegree{} being harder than either 5\textdegree{} or
15\textdegree{}.
Friction~$\mu$ shows a non-monotonic profile in which
intermediate friction (0.7) is the easiest condition (182~s),
with both low and high friction performing worse.
Target distance~$d$ produces a counter-intuitive inversion,
with the shortest range (0.2~m, 970~s) performing worst.
The heuristic setting ($k_{\text{rot}}=0.125$, $k_{\text{trans}}=0.3$,
$v_{\text{min}}=0.15$~m/s, $c=0.175$~m,
mean~$=245.5$~s) remains the best-performing configuration
across all environmental factor levels.

% ── Uphill terrain ────────────────────────────────────────
\subsection{Uphill Terrain (5\textdegree)}

Uphill terrain is the most demanding condition:
15 of 81 runs failed to complete within the time limit, and the
mean completion time among the 66 successful runs is 1256.4~s,
approximately 22$\times$ longer than downhill.

Among heuristic parameters (Fig.~\ref{fig:heuristic}, centre),
$v_{\text{min}}$ emerges as a relevant factor on uphill terrain
($\Delta = 465$~s), in contrast to flat and downhill conditions
where it has no effect.
The lowest tested value ($v_{\text{min}} = 0.10$~m/s) yields
the best performance (946~s); both higher levels produce
identical mean times of 1412~s.
Contact placement produces a non-monotonic profile ($\Delta = 923$~s):
the intermediate level ($c = 0.175$~m, 1715~s) performs worst,
while the highest level ($c = 0.300$~m, 792~s) is best---a
complete reversal from flat-terrain behaviour where $c = 0.300$~m
caused outright failures.
The gains $k_{\text{rot}}$ and $k_{\text{trans}}$ remain of
secondary importance ($\Delta \approx 98$~s each).

Target angle~$\theta$ is the dominant environment factor by a
large margin ($\Delta = 4820$~s, 56.3\% of environmental spread),
with a strongly non-monotonic profile (Fig.~\ref{fig:env}, centre).
The shallowest angle (5\textdegree) is by far the hardest condition
(5183~s mean), while 10\textdegree{} and 15\textdegree{} are
substantially easier (363~s and 404~s respectively).
Friction ($\Delta = 1485$~s, 17.3\%) is the second-largest
environmental factor, with a non-monotonic profile: the lowest
friction (0.4) is by far the easiest condition (223~s), while
intermediate and high friction are substantially harder (1708~s
and 1561~s respectively).
Target distance ($\Delta = 1239$~s, 14.5\%) shows a sharp
step from the shortest range (0.2~m, 360~s) to longer ranges,
with the 0.4~m and 0.6~m levels producing comparable and much
higher times (1599~s and 1584~s).
Mass ($\Delta = 1022$~s, 11.9\%) increases roughly monotonically
with box weight (568~s, 1418~s, 1590~s), as heavier boxes
require more pushing effort on the incline.
The best-performing uphill configuration ($k_{\text{rot}}=0.5$, $k_{\text{trans}}=0.3$,
$v_{\text{min}}=0.10$~m/s, $c=0.300$~m,
mean~$=368.1$~s) differs from the flat one on every
heuristic parameter.

% ── Downhill terrain ──────────────────────────────────────
\subsection{Downhill Terrain (5\textdegree)}

All 81 downhill runs completed successfully.
The mean completion time is 57.7~s, reflecting the gravitational
assist that makes this condition straightforward.

All heuristic parameters except contact placement produce
$\Delta = 0$~s (Fig.~\ref{fig:heuristic}, right), confirming
that the downhill task is insensitive to controller tuning within
the tested ranges.
Contact placement shows a small but monotonically beneficial
effect ($\Delta = 20$~s): higher contact placement ($c = 0.300$~m)
performs best (47~s), opposite to flat terrain.

The four environment factors contribute roughly balanced
influence (Fig.~\ref{fig:env}, right):
target distance ($\Delta = 74$~s, 30.3\%),
mass ($\Delta = 72$~s, 29.4\%),
friction ($\Delta = 59$~s, 24.3\%), and
target angle ($\Delta = 39$~s, 16.1\%).
Their profiles differ in shape.
Friction increases monotonically with coefficient: the lowest
friction (0.4) is easiest (22~s) and the highest (0.9) is hardest
(81~s), as greater surface resistance partially offsets the
gravitational assist.
Mass decreases monotonically: the lightest box (0.1~kg, 105~s)
is substantially harder than heavier boxes (35~s and 33~s),
since without the added gravitational load the robot loses
contact with the box more frequently.
Target distance increases monotonically, with the shortest
range (0.2~m, 15~s) completing far faster than longer ranges
(0.4~m: 69~s; 0.6~m: 89~s).
Target angle is the only non-monotonic factor (37~s, 77~s, 59~s),
with the intermediate level of 10\textdegree{} being hardest.
The best downhill configuration ($c=0.300$~m, with $k_{\text{rot}}$,
$k_{\text{trans}}$, and $v_{\text{min}}$ inconsequential, mean~$=46.6$~s)
is again distinct from both the flat and uphill optima.

% ── Cross-terrain comparison ──────────────────────────────
\subsection{Cross-Terrain Comparison}

Three findings emerge consistently from this sensitivity analyses across all terrain conditions.
First, the best-performing heuristics configuration is terrain-dependent:
values best-suited to the flat, uphill, and downhill scenarios varies across the heuristic
parameters; and no single controller setting simultaneously
performs the best across all three conditions.
Second, contact placement~$c$ is the only heuristic parameter
that produces a meaningful main effect on all three terrains,
though the nature (positive or negative) and magnitude of that effect differ sharply:
$c = 0.300$~m is catastrophic on flat terrain, best-performing on
downhill and uphill; while $c = 0.175$~m is
best-performing on flat but worst on uphill.
Third, the scale of environmental sensitivity varies dramatically
with slope condition:
the largest single environmental effect range is 1115~s on flat,
4820~s on uphill, and only 74~s on downhill,
spanning more than one order of magnitude across conditions.

\section{Physical experiments}\label{sec:hardware}

% \vspace{-3mm}

We use a lab-scale setup to implement and validate R2P2 in flat terrain. All the simulation experiments discussed in Section \ref{sec:num-expts} used 6 robots. The use of 6 robots is essential to ensure caging for up-hill and down-hill experiments. In contrast, in the case of flat-terrain transport, unlike caging-based methods (e.g. \cite{fink2008multi, wan2017multirobot, vardharajan2022collective}), R2P2 does not require a caging condition to achieve transport. In our physical experiments on flat terrain, we test our framework with a team of four robots. The experimental setup consists of TurtleBot3 Burger robots together with a rectangular cardboard-box measuring 1.2 m in length, 0.22 m in width, 0.21 m in height and weighing 1.2 kg. 
During preliminary trials, we noticed that robots sometimes lost traction while pushing the object. This caused a lack of force transfer and made it difficult for the robots to move the object effectively. To address this issue, we added 500 g mass to the front of each robot. This extra mass increased the normal force on the wheels, which improved ground traction and allowed more reliable object pushing.

\begin{figure*}[!ht]

\begin{minipage}[h]{0.48\linewidth}
    \centering

\includegraphics[width=0.95\linewidth]{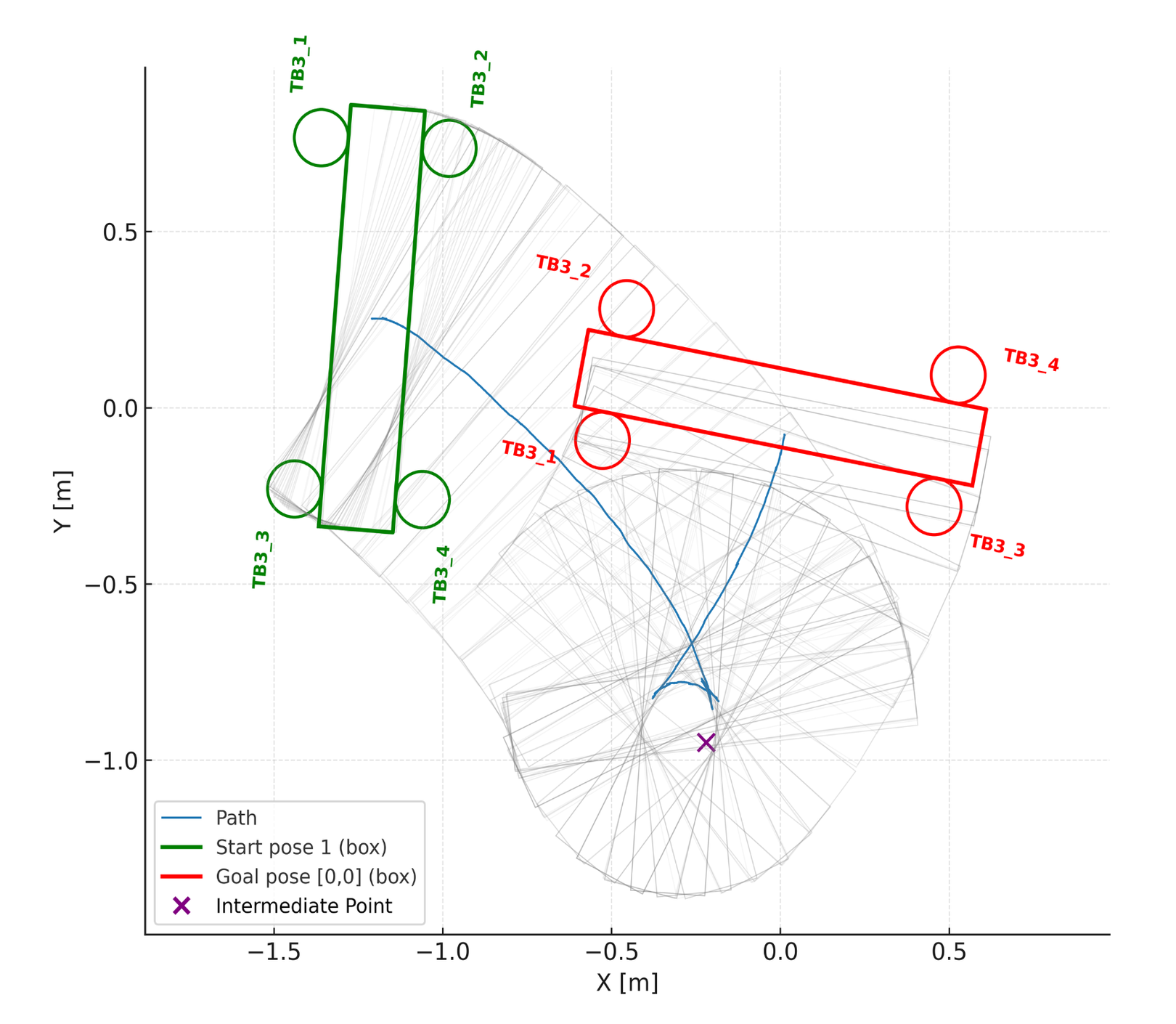}
\subcaption[]{Physical experiment box trajectory}
\label{subfig:phy-expt-box-traj}
\end{minipage}
\begin{minipage}[h]{0.48\linewidth}
    \centering

\includegraphics[width=0.92\linewidth]{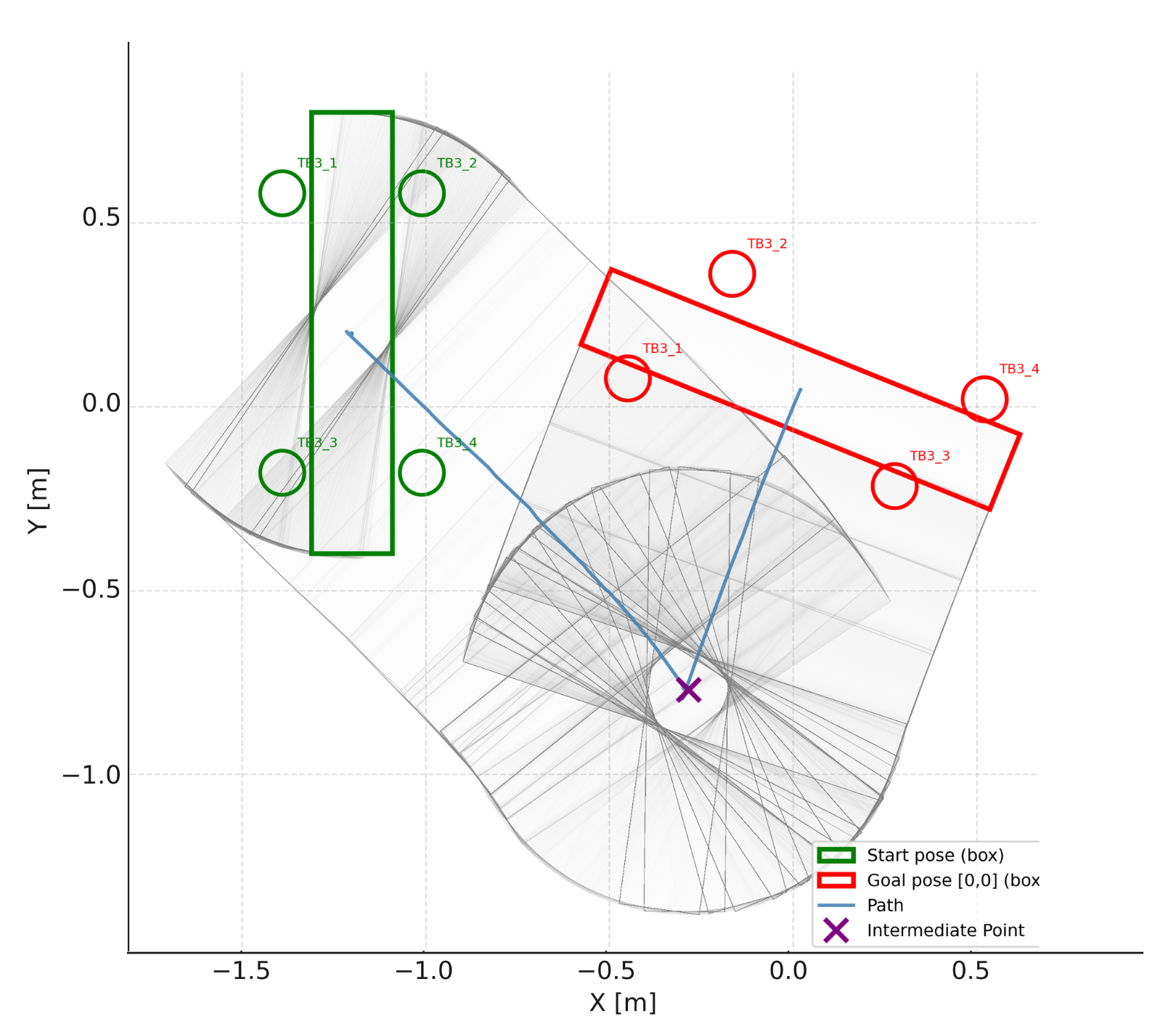}
\subcaption[]{Simulation experiment box trajectory}
% \begin{figure}
%     \centering
%     \includegraphics[width=1\linewidth]{image.png}
%     \caption{Enter Caption}
%     \label{fig:placeholder}
% \end{figure}
\label{subfig:sim-replica-box-traj}
\end{minipage}

\begin{minipage}[h]{0.24\linewidth}
    \centering
    \includegraphics[width=0.9\linewidth]{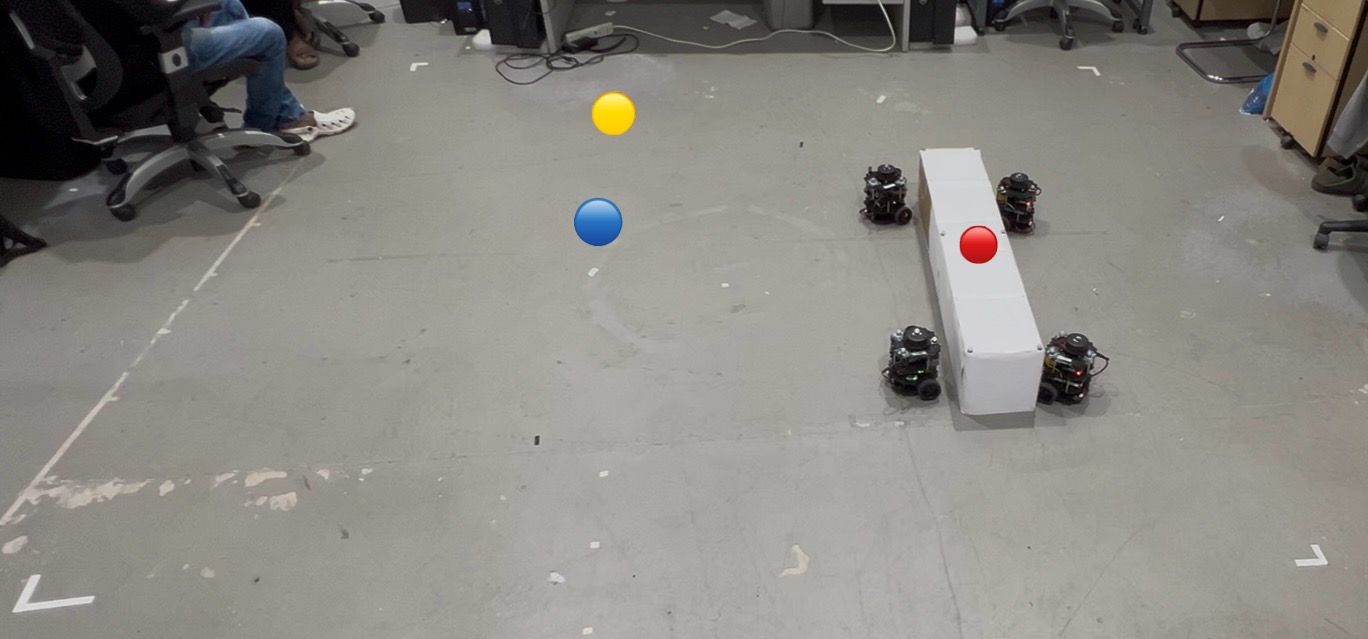}
    \subcaption[]{}
    \label{subfig:dwn-1-real}
\end{minipage}
\begin{minipage}[h]{0.24\linewidth}
    \centering

    \includegraphics[width=0.9\linewidth]{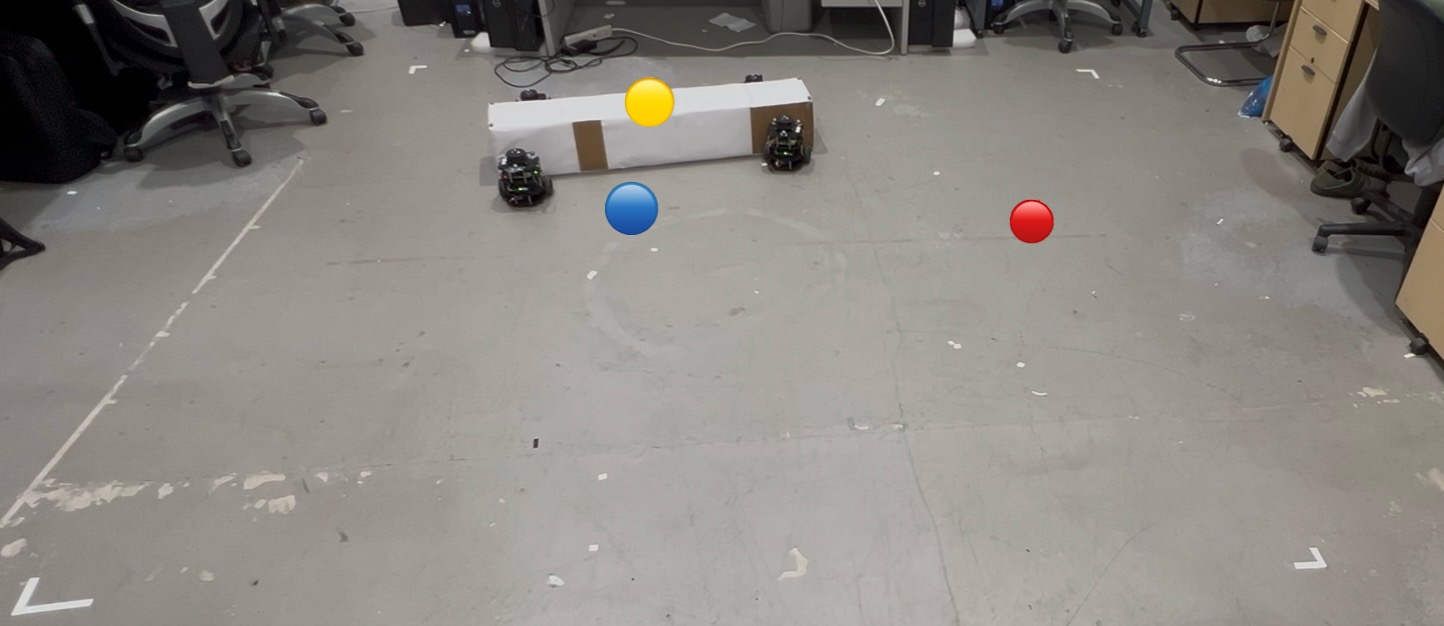}
    \subcaption[]{}
    \label{subfig:dwn-2-real}
\end{minipage}
\begin{minipage}[h]{0.24\linewidth}
    \centering

    \includegraphics[width=0.9\linewidth]{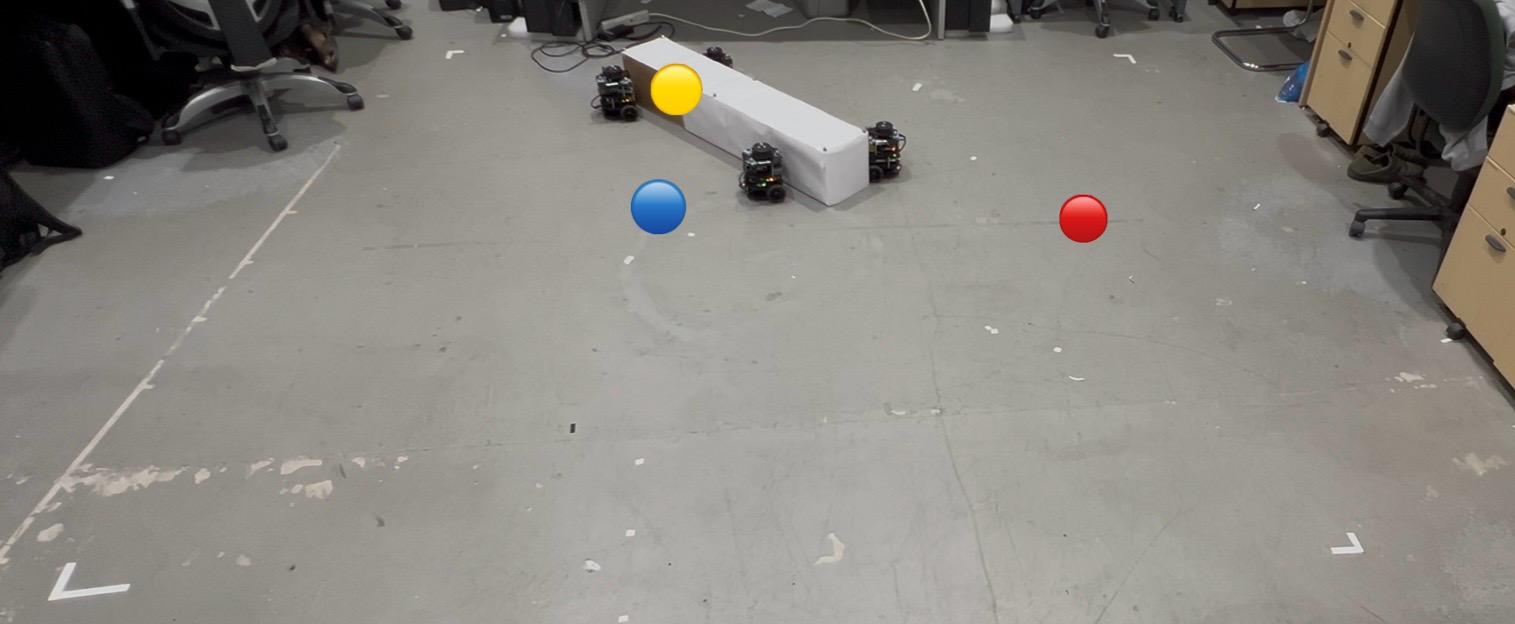}
    \subcaption[]{}
    \label{subfig:dwn-3-real}
\end{minipage}
\begin{minipage}[h]{0.24\linewidth}
    \centering

    \includegraphics[width=0.9\linewidth]{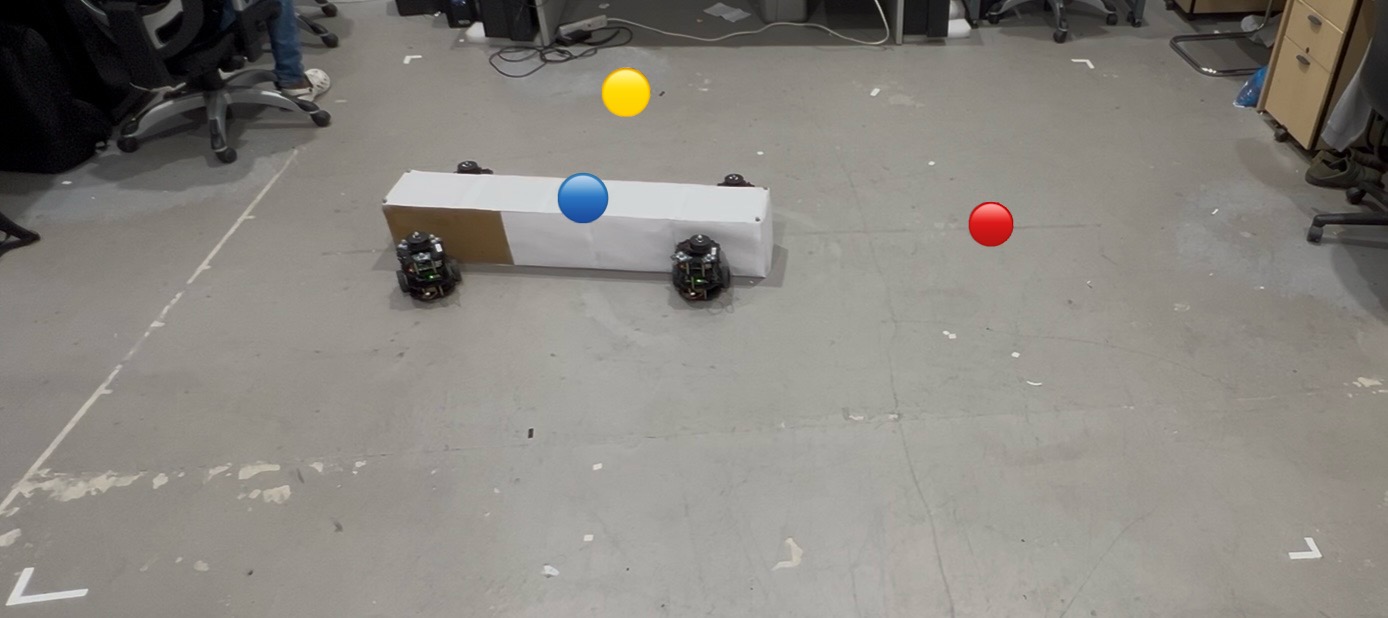}
    \subcaption[]{}
    \label{subfig:dwn-4-real}
\end{minipage}

\begin{minipage}[h]{0.24\linewidth}
    \centering

    \includegraphics[width=0.9\linewidth]{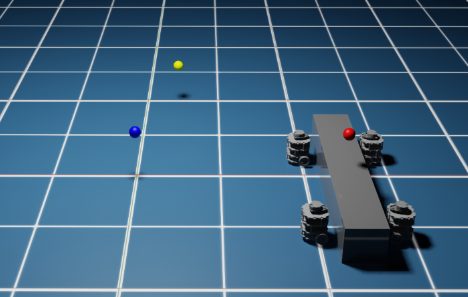}
    \subcaption[]{}
    \label{subfig:dwn-1-sim}
\end{minipage}
\begin{minipage}[h]{0.24\linewidth}
    \centering

    \includegraphics[width=0.9\linewidth]{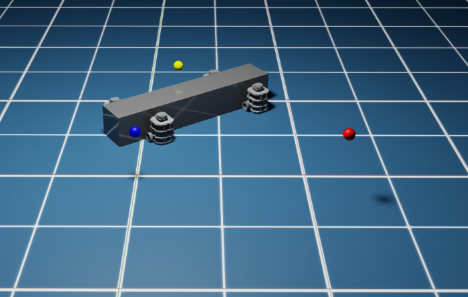}
    \subcaption[]{}
    \label{subfig:dwn-2-sim}
\end{minipage}
\begin{minipage}[h]{0.24\linewidth}
    \centering

    \includegraphics[width=0.9\linewidth]{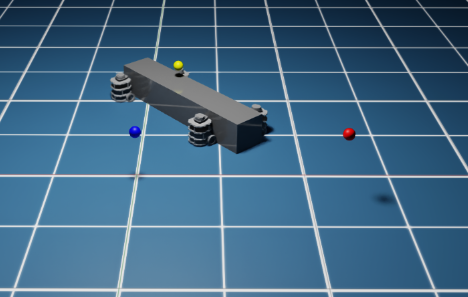}
    \subcaption[]{}
    \label{subfig:dwn-3-sim}
\end{minipage}
\begin{minipage}[h]{0.24\linewidth}
    \centering

    \includegraphics[width=0.9\linewidth]{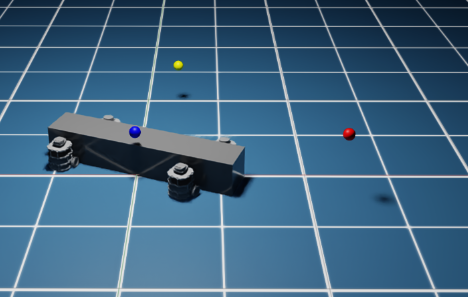}
    \subcaption[]{}
    \label{subfig:dwn-4-sim}
\end{minipage}
\caption{Physical and simulation experiment with R2P2. Plots (a) and (b) show the resultant box trajectory in the physical experiment and simulation respectively. Below them, Figure \ref{subfig:dwn-1-real} - \ref{subfig:dwn-4-real}  show snapshots over the physical experiment. Figure \ref{subfig:dwn-1-real} shows the initial box state, \ref{subfig:dwn-2-real} shows the box reached the first waypoint, in figure \ref{subfig:dwn-3-real} the box is rotated at the first waypoint towards the goal position and in figure \ref{subfig:dwn-4-real} the box reached its goal position. The corresponding simulation snapshots are shown in figures \ref{subfig:dwn-1-sim} - \ref{subfig:dwn-4-sim}. 
% \textcolor{red}{Plots (a) and (b) needs to be of the exact same size, with same x-y axis}. 
}  
\label{fig:sim-hardware-same-expt}
\end{figure*}

Each robot is equipped with Raspberry Pi 4 Model B. We use a Motion Capture (MoCap) system to get the position and orientation of all the robots and the box. The data from the MoCap is streamed at a rate of 40 Hz. 
The entire software pipeline is implemented using ROS2.
We implemented R2P2 in a decentralized fashion, with each robot making its own decisions on its on-board processor. The software architecture diagram will be released along with the software package as a public repository once the paper is accepted. 

Figure \ref{subfig:phy-expt-box-traj} shows the resulting box trajectory for this experiment. The object is initially placed at ( -1.2, 0.2) in the global frame. We set the goal point at (0, 0) and an intermediate waypoint at (-0.22, -0.95). All coordinates are in meters. The box is turned by an angle of about 40 degrees (clockwise) by the robots at its initial position to orient its heading towards the intermediate waypoint. Then, the box is linearly translated to the intermediate waypoint. At that point, the object is rotated about 120 degrees anti-clockwise, and then translated linearly to the goal; the goal is reached successfully within the tolerance defined (0.1 m). The mission is accomplished in about 25 minutes. The snapshots of the real-world experiment are presented in Fig.~\ref{subfig:dwn-1-real} - Fig.~\ref{subfig:dwn-4-real}. Figure ~\ref{subfig:dwn-1-real} depicts the initial position of the object with red dot. Figure~\ref{subfig:dwn-2-real} shows the intermediate position with yellow dot. Figure~\ref{subfig:dwn-3-real} illustrates the object's rotation after reaching the first goal. Figure~\ref{subfig:dwn-4-real} shows the final position of the object with blue dot. The video of the physical experiment is available here\footnote{https://buffalo.box.com/s/rpvqph1nke4vfqp3cxslnoytzm74zjbp}, as well as at the following Github repository \cite{shah_thota_r2p2_demo} (which also provides simulation experiment videos).

To assess how the physical experiment might differ from the experiment in simulation on which we performed our detailed evaluations presented earlier, we run the same experiment in our virtual environment. In simulation, the mission takes about 3 mins. Figure \ref{fig:sim-hardware-same-expt} shows the simulation experiment for the same initial box state, intermediate waypoint and the goal-location, as used in the physical experiment. Figure \ref{subfig:sim-replica-box-traj} shows the resulting box trajectory in the simulation, while Fig.~\ref{subfig:dwn-1-sim} to \ref{subfig:dwn-4-sim} show snapshots of the mission in the simulator. We use the exact same R2P2 heuristics in both the simulation and the physical experiment. 
A comparison of Fig.~\ref{subfig:phy-expt-box-traj} and Fig.\ref{subfig:sim-replica-box-traj} reveal that the box trajectory in simulation and the physical experiment are very similar. 
We do observe minor differences because of the robot-hardware and terrain related sim-to-real gaps. For instance, the straight motion in the simulation results in a straight line of the box-centroid while a bit of curvature is observed in the physical experiment box-trajectory. Figure~\ref{subfig:sim-replica-box-traj} shows that robots are able to achieve pure box-rotation at the intermediate waypoint in simulation; while in the physical experiment (Fig.~\ref{subfig:phy-expt-box-traj}), the box centroid undergoes some drift during rotation, i.e., a bit of unintended translation along with rotation. One favorably interesting observation was that in spite of loss of traction at times in the physical experiment, the rule based roles and control was able to adapt to provide successful mission completion, albeit at a notably larger physical time. In the future, domain randomization type approaches could be used to add stochasticity to the simulation environment parameters, to better match the transport behavior observed in simulation vs. the real environment. 

\section{Conclusion} \label{sec:conc}
In this paper, we developed a decentralized task and motion planning approach, \textbf{R2P2}, to collaboratively transport a box along a given reference path with multiple wheeled robots. 
After allocation of the initial robot positions, the robots decide their roles based on their relative position and the desired box maneuver. 
Specific to the role, each robot takes velocity decisions using a set of rules or proportional control.
We tested the R2P2 in an IsaacSim-based simulator. A team of six differential drive turtlebots were used to carry out the transport mission over a sequence of reference wapoints in up-hill, down-hill and flat terrain scenarios with multiple surface friction and box mass settings. R2P2 demonstrated successful transport for all the case-studies (reaching within a specified threshold distance of the goal point), showcasing its generalizability across the surface and box mass factors. While surface friction was not observed to have a clear monotonic impact on mission time of the transport, the latter was observed to linearly increase with the box mass. 
% \textcolor{red}{Lastly, scalability of the method with increasing team size was also demonstrated. }
We also performed a successful physical validation of R2P2 deployed onboard 4 TurtelBot3 Burger robots, that are tasked to move a rectangular cardboard box over a flat terrain. The robots were able to transport the object to the goal location through a specified intermediate waypoint.
% The heuristics of R2P2 were not tuned for the physical experiments
We conducted the same experiment in simulation, with the same heuristics as that used in the physical experiment. 
The box trajectories in simulation and the physical experiment were found to be similar. Minor differences in trajectory and notable difference in the recorded mission time was due to the robot-hardware and terrain related artifacts. Sensitivity analyses over the R2P2 heuristics and environmental factors illustrated that the contact placement is the dominant heuristic factor across all the terrains, e.g., showing 9 $\times$ performance difference on flat-terrains, 3.7 $\times$ on uphill scenarios and 2.2 $\times$ on down-hill scenarios. 
% The resultant box trajectory in simulation and ha with R2P2 over the same mission

Even though the R2P2 achieves successful transport over varying terrains, box and environment parameters, there's potential to improve the mission performance with respect to objectives such as overall mission time, energy cost or deviation from reference trajectory. The future extension of this work will thus explore both scenario specific optimization of the R2P2 heuristics (need for which was seen from the sensitivity analyses) and/or learning of these heuristics to generalize across scenarios in terms of surface properties, and box shape and mass. This will allow us to also explore the viability of the assumed observation space or deal with partial/uncertain observation issues to further enhance the application potential of this method.

\bibliographystyle{ieeetr}  %% .bst file following ASME conference format. Do not change.
\bibliography{references}%% <=== change this to the name of your bib file
\end{document}